\documentclass[12pt]{article}
\newcommand{\pmb}[1]{{\setbox0=\hbox{#1}%
		\kern-.025em\copy0\kern-\wd0
		\kern.05em\copy0\kern-\wd0
		\kern-.025em\raise.0433em\box0 }}
\newcommand{\fg}[1]{\mbox{\pmb{$#1$}}}
\usepackage{amssymb}
\usepackage{amsmath}
 
\usepackage{multirow}
\usepackage{latexsym}
\usepackage[pdftex]{graphicx}
\DeclareGraphicsExtensions{.pdf,.jpg}
\usepackage{epsf}
\usepackage{graphicx,epsfig}
\usepackage{epstopdf}
\usepackage{amssymb}
\usepackage{amsmath}
\usepackage{subfigure}
\usepackage{setspace}
\usepackage{color}
\usepackage{latexsym}
\usepackage{comment}
\usepackage[pdftex]{graphicx}
\usepackage[colorlinks,citecolor=blue]{hyperref}
\DeclareGraphicsExtensions{.pdf,.jpg}
\usepackage{multirow}
\usepackage{epsf}
\usepackage{multirow}
\usepackage[ruled,vlined]{algorithm2e}
\usepackage{array}
\usepackage{booktabs}
\usepackage{authblk}
\usepackage[none]{hyphenat}
\usepackage{microtype}
\DisableLigatures{encoding = *, family = *}
\usepackage{tikz}
 
\usepackage{pifont}% http://ctan.org/pkg/pifont
\usepackage{subfloat}
\usepackage{dblfloatfix}
\usepackage{epstopdf}
\usepackage{float}
\usepackage[round]{natbib} %Authors name et al. with round bracket \citep{}, \cite{} to 
\usepackage{setspace}
\newcommand{\myb}[1]{\mbox{\boldmath $#1$}}      %bold math font
\def\tenp{\myb{\otimes}}                         %tensor product symbol
\makeatletter
\newcommand*{\rom}[1]{\expandafter\@slowromancap\romannumeral #1@}
\makeatother
\newcommand{\bey}{\begin{eqnarray}}
	\newcommand{\eey}{\end{eqnarray}}

\newcommand{\bec}{\begin{center}}
	\newcommand{\eec}{\end{center}}

\begin{document}
	\pagestyle{myheadings}
	\setcounter{tocdepth}{2}
	\baselineskip22pt
	\belowdisplayskip11pt
	\belowdisplayshortskip11pt
	\renewcommand{\thefootnote}{\fnsymbol{hello}}
	\author[1]{Arunabha  M. Roy}%\thanks{Corresponding author, email address: arunabhr@umich.edu}}
\author[2]{Jayabrata Bhaduri}
\affil[1]{\it Aerospace Engineering Department, University of Michigan, Ann Arbor, MI 48109, USA}
\affil[2]{\it Capacloud AI, Deep Learning $\&$ Data Science Division, Kolkata, WB 711103, India.}
\title{\large \bf A Computer Vision  Enabled  damage detection model with  improved YOLOv5 based on Transformer Prediction Head }
\date{}
\maketitle
\noindent
\bec
{\bf Abstract}
\eec
{\it Objective.} 
Computer vision-based up-to-date accurate damage classification and localization are of decisive importance for infrastructure monitoring, safety, and the serviceability of civil infrastructure. Current state-of-the-art deep learning (DL)-based damage detection models, however, often lack superior feature extraction capability in complex and noisy environments, limiting the development of accurate and reliable object distinction.
{\it Method.} To this end, we present DenseSPH-YOLOv5, a real-time DL-based high-performance damage detection model where DenseNet blocks have been integrated with the backbone to improve in preserving and reusing critical feature information. 
Additionally, convolutional block attention modules  (CBAM) have been implemented to improve attention performance mechanisms for strong and discriminating deep spatial feature extraction that results in superior detection under various challenging environments. 
Moreover, an additional
feature fusion layers and a Swin-Transformer Prediction Head (SPH)  have been added leveraging advanced self-attention mechanism for more efficient detection of multiscale object sizes and simultaneously reducing the computational complexity.
{\it Results.} Evaluating the model performance in large-scale Road Damage Dataset (RDD-2018),  at a detection rate of 62.4 FPS,  DenseSPH-YOLOv5 obtains a mean average precision (mAP) value of $85.25 \%$, F1-score of $81.18 \%$, and precision (P) value of $89.51 \%$  outperforming current state-of-the-art models. 
{\it Significance.}  The present research provides an effective and efficient damage localization model addressing the shortcoming of existing DL-based damage detection models by providing highly accurate localized bounding box prediction. Current work constitutes a step towards an accurate and robust automated damage detection system in real-time in-field applications.
\\
\\
{\bf 1. Introduction :}
\\
\\
In recent years,  automated damage detection plays an  important  role in various industrial applications including 
product quality assessment  \citep{agarwal2015adaptive,hanzaei2017automatic}, infrastructure monitoring \citep{eisenbach2017get,gopalakrishnan2018deep}, safety and the serviceability of civil infrastructure \citep{koch2015review,hartmann2020advanced}. 
Early-stage accurate crack detection is critical for pavement
damage rating and subsequent sealing or rehabilitation activities \citep{chen2021pavement,chen2022crackembed}.
Therefore, it is important for roadway infrastructure engineers to detect pavement
cracks accurately so that the best cost-effective plans of maintenance
and rehabilitation could be employed \citep{ni2022generative,dong2021framework}.
While traditional damage detection techniques mainly include visual inspection, however, such labor-intensive methods have disadvantages due to their low efficiency, high cost, and individual biases \citep{xu2022efficient,shang2023defect}.
Additionally, it is also limited in reproducibility, reliability, and objectivity due to the requirement of qualified personnel for domain-specific experience,  knowledge, and skill sets \citep{fang2020novel}.
To circumvent such issues, more recently,  various automatic and semi-automatic crack detection algorithms have gained significant attraction \citep{gopalakrishnan2018deep}.

For the last three decades,  image-based crack detection \citep{mohan2018crack,koch2015review} that include various image processing approaches, such as edge detection \citep{zhao2010improvement,hanzaei2017automatic,li2022method}, dynamic thresholding \citep{oliveira2009automatic, wang2021crack}, and different morphological operations  \citep{anitha2021survey,li2021pixel} have been the central focus for detecting damage in challenging real-world scenarios.  However, the aforementioned methods are quite sensitive to noise and varying illumination intensities, and hence, not suitable in real-world complex conditions \citep{koch2015review}. 
To circumvent such issues, later, conventional machine learning (ML) approaches have been introduced for damage detection  \citep{mohan2018crack,koch2015review}.
In general, such methods utilize a trained classifier such as a Support Vector Machine (SVM) on local feature descriptors that can be either Local Binary Patterns (LBP) \citep{varadharajan2014vision,quintana2015simplified} or Histogram of Oriented Gradient (HOG) \citep{kapela2015asphalt}. 
Although, compared to conventional image processing approaches,  ML-based models significantly improve the accuracy and efficiency of the
damage detection, however,  due to large errors in classification performances remains a serious bottleneck for deploying such models in real-world applications \citep{fang2020novel}.

%==========================DL ================================

More recently, deep learning (DL) characterized by multilayer neural networks (NN)  \citep{lecun2015deep}
has shown  remarkable  breakthroughs in pattern recognition for various fields including
image classification \citep{rawat2017deep,jamil2022distinguishing,khan2022introducing,khan2022sql}, computer vision \citep{voulodimos2018deep,roy2021deep,roy2022fast,roy2022real,roy2022computer}, 
object detection \citep{zhao2019object,chandio2022precise,roy2022wildect,singh2023deep}, 
brain-computer interfaces \citep{roy2022efficient,roy2022adaptive,roy2022multi,singh2023understanding}, signal classification \citep{jamil2023efficient,jamil2023robust} 
and across diverse scientific disciplines  \citep{bose2022accurate,roy2023physics,royguha2022,roybose2023,royguha2023}.
%In recent years, driven by the advancement of bigdata-based  architectures \citep{khan2022sql},  deep  learning (DL) techniques \citep{lecun2015deep} have shown great promises in computer vision \citep{voulodimos2018deep,roy2021deep,roy2022fast,roy2022real,roy2022computer}, object detection \citep{zhao2019object,chandio2022precise,roy2022wildect,singh2023deep}, image classification \citep{rawat2017deep,irfan2021role,jamil2022distinguishing,khan2022introducing}, damage detection \citep{guo2022damage,glowacz2022thermographic,glowacz2021fault} 
%brain-computer interfaces \citep{roy2022efficient,roy2022adaptive,roy2022multi,singh2023understanding} and  
%across various scientific applications
%\citep{butler2018machine,ching2018opportunities,bose2022accurate}.
Following the success, there is an increasing thrust of research works geared towards damage classification tasks employing DL techniques, mostly convolutional neural networks (CNN), such as ResNet \citep{bang2018deep}, AlexNet \citep{dorafshan2018sdnet2018,li2018pixel}, VGG-net \citep{gopalakrishnan2017deep,silva2018concrete} and various others \citep{chow2020anomaly,nath2022drone,li2021automatic}.  
Particularly in object localization,  DL methods 
have demonstrated superior accuracy  \citep{Han_et_al-2018} that can be categorized into two classes: two-stage and one-stage detector \citep{Lin_et_al-2017}.
Two-stage detectors including  Region Convolution Neural Network (R-CNN) \citep{Girshick-2015}, faster R-CNN  \citep{Ren_et_al-2015}, mask R-CNN \citep{He_et_al-2017} etc that have shown a significant improvement in accuracy in object localization. 
In recent times,  You Only Look Once (YOLO) variants  \citep{Redmon_et_all-2016,Redmon_Farhadi-2017,Redmon_et_all-2018,Bochkovskiy_et_all-2020} have been proposed that unify target classification and localization. In \citep{roy2022fast,roy2022real,roy2021deep}, DenseNet \citep{Huang-IEEE-2017}  
blocks attaching  Spatial Pyramid Pooling (SPP)  \citep{He-IEEE-2015} with an improved modified Path Aggregation Network (PANet) \citep{Liu-IEEE-2018} has been integrated to enhance the representation of receptive fields and extraction of important contextual features in the original YOLOv4 leading to significant improvement in the detection speed and accuracy. 
In order to enhance gradient performance and reduce the computational cost, YOLOv4 \citep{Bochkovskiy_et_all-2020} designs a cross-stage partial (CSP) network. To further improve detection accuracy, YOLOv4 implements Mish activation  \citep{Misra-2019} and CIoU loss \citep{Zheng_et_all-2020}.
Recently,   Scaled-YOLOv4 \citep{wang2021scaled} demonstrated its superior performance in detecting the vast range of linearly scaled objects for various applications. As the latest generation of
the YOLO series, the YOLOv5 \citep{jocher2021ultralytics} has been rated top among state-of-the-art object detection models which inherit all the above advantages. Thus, in the present work, YOLOv5 has been considered a benchmark model for multiclass damage detection. 
More recently, improved YOLOv5 Based on Transformer Prediction Head (TPH-YOLOv5) \citep{zhu2021tph} has been proposed integrating convolutional block attention module (CBAM) \citep{woo2018cbam} for closely packed object detection and 
Swin Transformer-enabled YOLOv5 (SPH-YOLOv5) \citep{gong2022swin} has been designed incorporating Normalization-based Attention Modules
(NAM) that demonstrate significant improvement in accuracy while simultaneously reducing the computational complexity of the model which are the main motivations for the network architectural development of the current work. 
\\
\\
{\bf 1.1  Related Works :}
\\
\\
In this section, some recent and relevant DL works have been highlighted in the field of road damage detection.  In recent years, multiple studies have adopted various ML and DL-based approaches for automated road surface damage classification and detection  \citep{zhang2017automated,stricker2019improving,biccici2021approach}
For instance, a smartphone-based supervised deep convolutional
neural network (D-CNN) has been proposed for pavement damage classification \citep{zhang2016road}. Along a similar line, deep neural network (DNN) architecture has been employed for detecting cracks and potholes \citep{anand2018crack, silva2018concrete} as well as pavement condition assessment \citep{fan2018automatic}.  
In \cite{nhat2018automatic}, the superiority of the DCNN-based approach has been demonstrated over edge-detection-based approaches for pavement crack
detection. In \cite{maeda2018road}, a real-time road damage detection model based on SDD has been proposed that has been trained on a publicly available large-scale road damage dataset (RDD-2018) for eight different categories of road damages. Due to the popularity of the dataset, various attempts have been made,  notably using  YOLO \citep{alfarrarjeh2018deep}, Faster R-CNN 	\citep{kluger2018region},  Faster R-CNN with ResNet-152 \citep{wang2018road}, Ensemble models with Faster R-CNN and SSD \citep{wang2018deep}, and RetinaNet \citep{angulo2019road} to further improve the detection performance.
In addition, RetinaNet \citep{angulo2019road} has been used on a modified RDD-2018 dataset which demonstrates significant performance improvement. 
Following the work of  \cite{maeda2018road}, progressive growing- generative adversarial networks (PG-GANs) \citep{maeda2021generative} with Poisson blending have been used to generate new training data (RDD-2019) in order to  improve the accuracy
of road damage detection. 
More recently, transfer learning (TL)-based road damage detection model \citep{arya2021deep} has been proposed introducing large-scale open-source dataset RDD2020 \citep{arya2020global,arya2021rdd2020} considering multiple countries. 
In \cite{naddaf2020efficient}, EfficientDet-D7 has been employed for the detection of asphalt pavement distress. Whereas, YOLO CSPDarknet53 	\citep{mandal2020deep} has been used for road damage detection. Similarly, the YOLO network has been used for detecting pavement distress from high-resolution images \citep{du2021pavement}.  In 	\cite{majidifard2020pavement}, YOLOv2 model has been utilized for pavement distress classification from Google street view images. Along a similar line, a CNN-based predictive model trained in  Google API images has been employed for detecting potholes in   \cite{patra2021potspot}.
In a separate work in \cite{guan2021automated}, a stereo-vision integrated segmentation-based DL model with  modified depth-wise
separable convolution  U-net has been deployed for crack and pothole detection where multi-feature image datasets have been used to train the model. 
More recently,  a semi-supervised DL-based pixel-level segmentation model  \citep{karaaslan2021attention} has been proposed utilizing attention guidance for cracks and spalls localization that reduces computational cost significantly. 	
In separate work, an improved YOLOv5 road damage detection algorithm \citep{guo2022road} has been proposed leveraging MobileNetV3 as a backbone feature extractor. In \cite{haciefendiouglu2022concrete}, Faster R-CNN has been employed for concrete pavement crack detection under various illumination and weather conditions. Although, there exist several state-of-the-art works for damage detection 
including multi-class damage localization models, however, they often suffer from low accuracy, missed detection, and relatively large computational overhead \citep{cao2020survey,azimi2020data,naddaf2020efficient}.
\\
\\	
{\bf 1.2 Motivations :}
\\
\\
Despite illustrating outstanding performance in damage detection, current state-of-the-art DL algorithms still require further improvement due to their insufficient fine-grain contextual feature extraction capability leading to missed detection and false object predictions for various damages/cracks which possess a wide range of textures, shapes, sizes, and colors \citep{cao2020survey,azimi2020data,naddaf2020efficient}. 
Between various damage classes, accurate detection and localization tasks can be challenging due to significant variability of lightening conditions, low visibility, the coexistence of multi-object classes with various aspect ratios, and other morphological characteristics \citep{azimi2020data,naddaf2020efficient}.  
Additionally,  visual similarities, complex backgrounds,  and various other critical factors offer additional difficulties for the state-of-the-art damage detection models \citep{naddaf2020efficient}. 
%Thus, the main motivation of the present study is to design an efficient and robust computer vision-based algorithm for the accurate classification and localization of various crack/damage detection. Moreover, 	tiny objects with irregular intensity and shape variations as well as noisy and complex backgrounds induce additional challenges \citep{fang2020novel,naddaf2020efficient}. 
To this end, the current works aim to develop an efficient and robust damage classification and accurate damage localization model simultaneously productive in terms of training time and computational cost which is currently lacking in the recent state-of-the-art endeavors. 
\\
\\		
{\bf 1.3 Contributions :} 
\\
\\
To address the aforementioned shortcomings, in the current study,  we present DenseSPH-YOLOv5  based on an improved version of the state-of-art YOLOv5  detection model for accurate real-time damage detection.
The major contributions of the present research work can be summarized as follows:

\begin{itemize}
	
	\item In Dense-SPH-YOLOv5,  we have attached    DenseNet 
	%\citep{Huang-IEEE-2017}  
	blocks with CSP modules  in the CSPDarknet53  to
	preserve critical feature maps and efficiently reuse the discriminating feature information.
	
	\item 	Secondly,  we have introduced an additional detection head specifically for detecting tiny objects in the head part of the proposed DenseSPH-YOLOv5 network.

	\item 	In addition, the convolutional block attention module (CBAM) 
	%\citep{woo2018cbam} 
	has been implemented  for the construction of  progressive feature attention with large coverage along both channel and spatial 
	dimensions for strong and discriminating deep spatial feature extraction during object detection. 
	
	\item Furthermore, the regular CNN prediction heads (CPH)  in YOLOv5 have been upgraded with  Swin transformer Prediction Heads (SPHs) employing Swin transformer (STR) encoder block % \citep{liu2021swin} 
	leveraging advanced self-attention mechanisms for efficient detection of multi-scale object sizes and simultaneously reducing the computational complexity. 
	
	\item Spatial Pyramid Pooling (SPP)  
	%\citep{He-IEEE-2015} 
	has been tightly attached to the backbone to enhance the representation of receptive fields and extraction of important contextual features.
	
	\item Finally, an improved modified Path Aggregation Network (PANet) %\citep{Liu-IEEE-2018} 
	has been utilized to efficiently preserve fine-grain localized information by feature fusion in a multi-scale feature pyramid map. 
	
\end{itemize}

\begin{figure}
	\noindent
	\centering
	\includegraphics[width=0.7\linewidth]{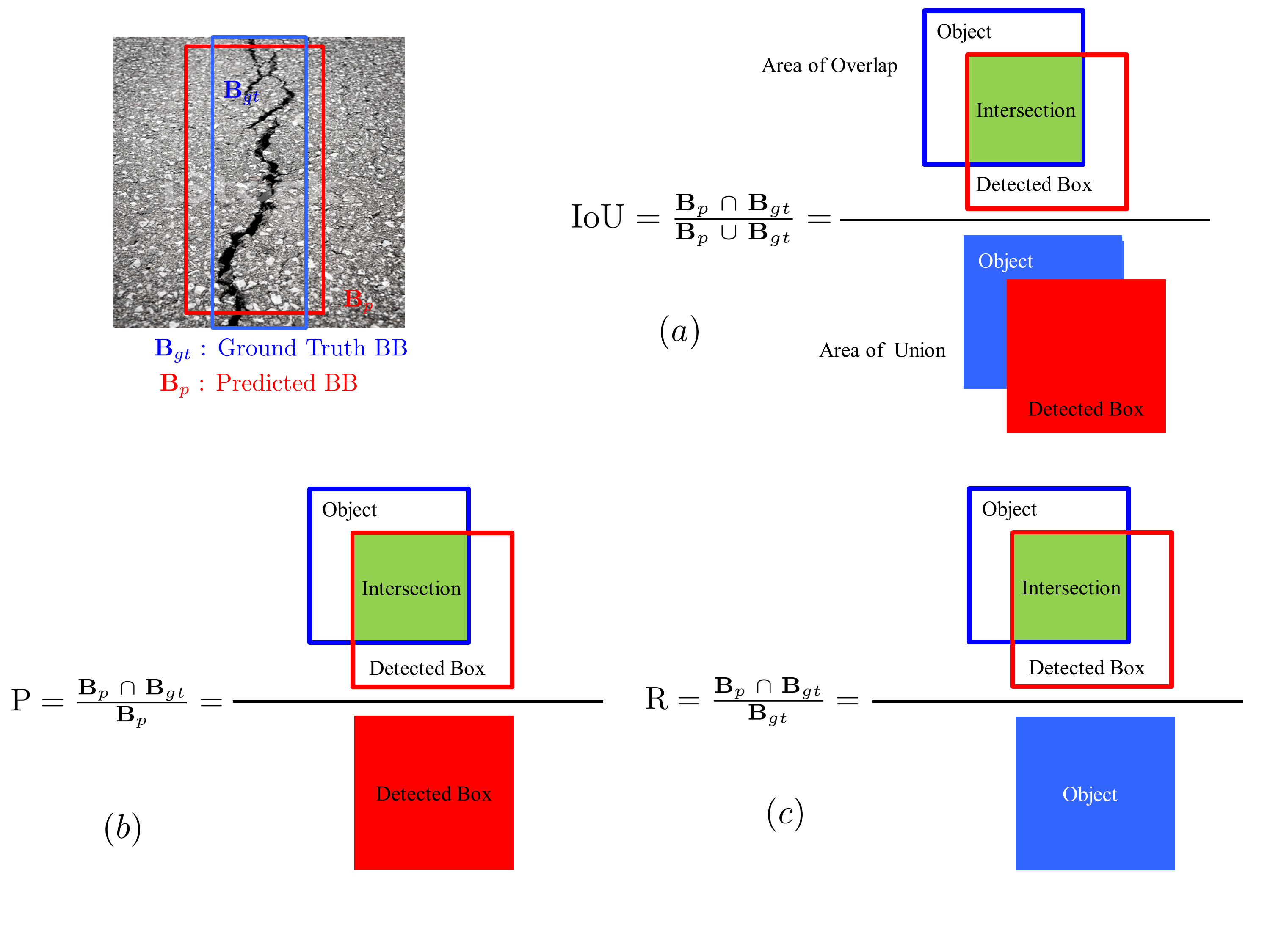}
	\caption{\label{Fig-2} Schematic illustration of measuring various performance metrics: (a)  Intersection over Union (IoU); (b) precision (P); (c) recall (R) during damage detection process.  }
\end{figure}

\begin{figure}[h]
	\centering
	\includegraphics[width=1.1\textwidth]{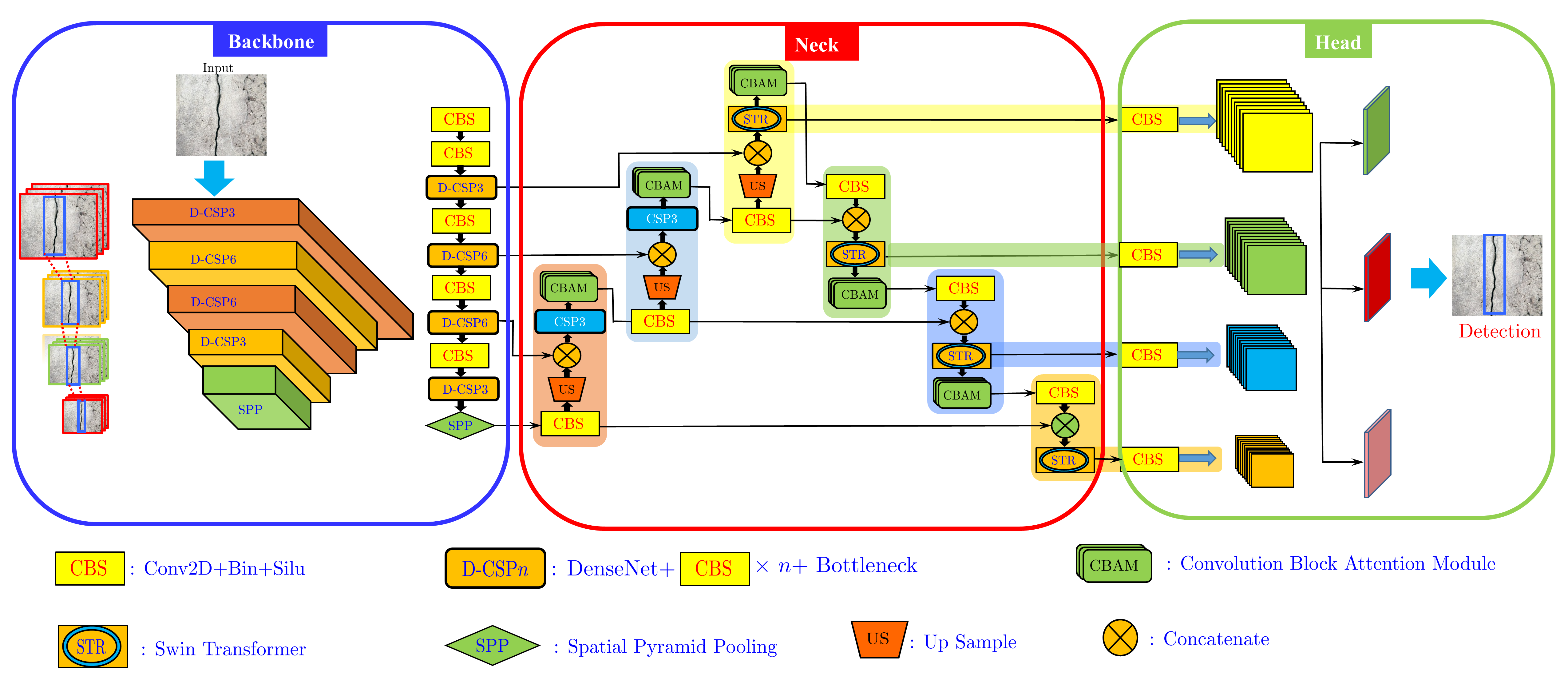}
	\caption{Schematic of the proposed DenseSPH YOLOv5 with fused DenseNet in CSPDarknet53 backbone; CBAM module in the improved neck; 
		four SPHs utilize hierarchical feature maps from STR encoder blocks in the modified Neck. }
	\label{Fig-3}
\end{figure}

\begin{figure}
	\centering
	\includegraphics[width=1\textwidth]{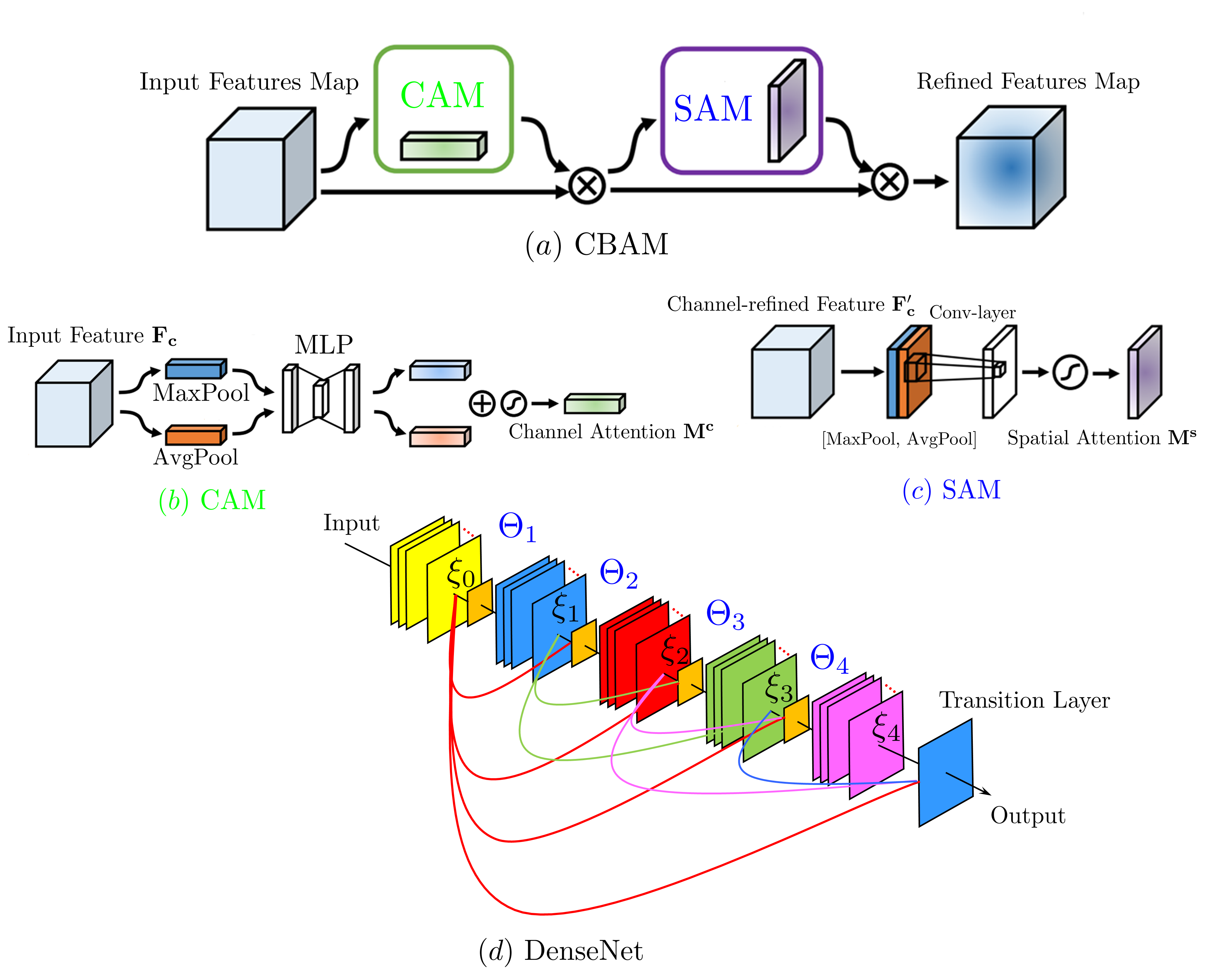}
	\caption{Schematic of (a) CBAM that has two sequential sub-modules:  (b) CAM and (c) SAM for adaptive refinement of  feature map at every intermediate convolution  blocks; (d) network architecture of  DenseNet  block  used in  Dense-SPH YOLOv5 detection model.}
	\label{Fig-4}
\end{figure}

With the aforementioned modifications, the detection accuracy of the model has been significantly enhanced for multi-scale object detection. An extensive ablation study has been performed for different combinations of backbone-neck architecture in order to optimize both accuracies of detection and detection speed. 
The proposed DenseSPH-YOLOv5 has been employed to detect distinct eight different damage classes that provide superior and accurate detection under various complex and challenging environments.  
%Evaluated on a custom-made dataset for various crack detection, it has been  found that  at a detection rate of 59.20 FPS,     Dense-SPH-YOLOv5 has achieved mean-average precision (mAP), F1, and precision (P) values of  $96.89 \%$,  $97.87 \%$, and  $97.18 \%$, respectively outperforms existing state-of-the-art crack detection models in terms of both classification accuracy and localized bounding box prediction  in detecting all crack classes.
The present work effectively addresses the shortcoming of existing DL-based crack detection models and illustrates its superior potential in real-time in-field applications. 
%In short, current work constitutes a step toward a  fully automated and efficient  DL-based  computer vision  methodology for multi-class   damage  detection processes for in-field applications. 
The rest of the paper is organized as follows: description of the dataset has been described in Section 2; Section 3 introduces the proposed methodology for damage detection;  the relevant finding and discussion of the proposed model have been discussed in Sections 4 and 5, respectively. Finally, the conclusions of the present work have been discussed in section 6.
\\
\\
{\bf 2. Road Damage Dataset :}
\\
\\
In the current study, large-scale Road Damage Dataset (RDD-2018)  \citep{maeda2018road} has been used that consists of 9,053 labeled road damage
images of resolution $600 \times 600$ pixels containing a total number of 15,435 annotated bounding boxes for eight different types of road damage.
\begin{figure}
	\centering
	\includegraphics[width=\textwidth]{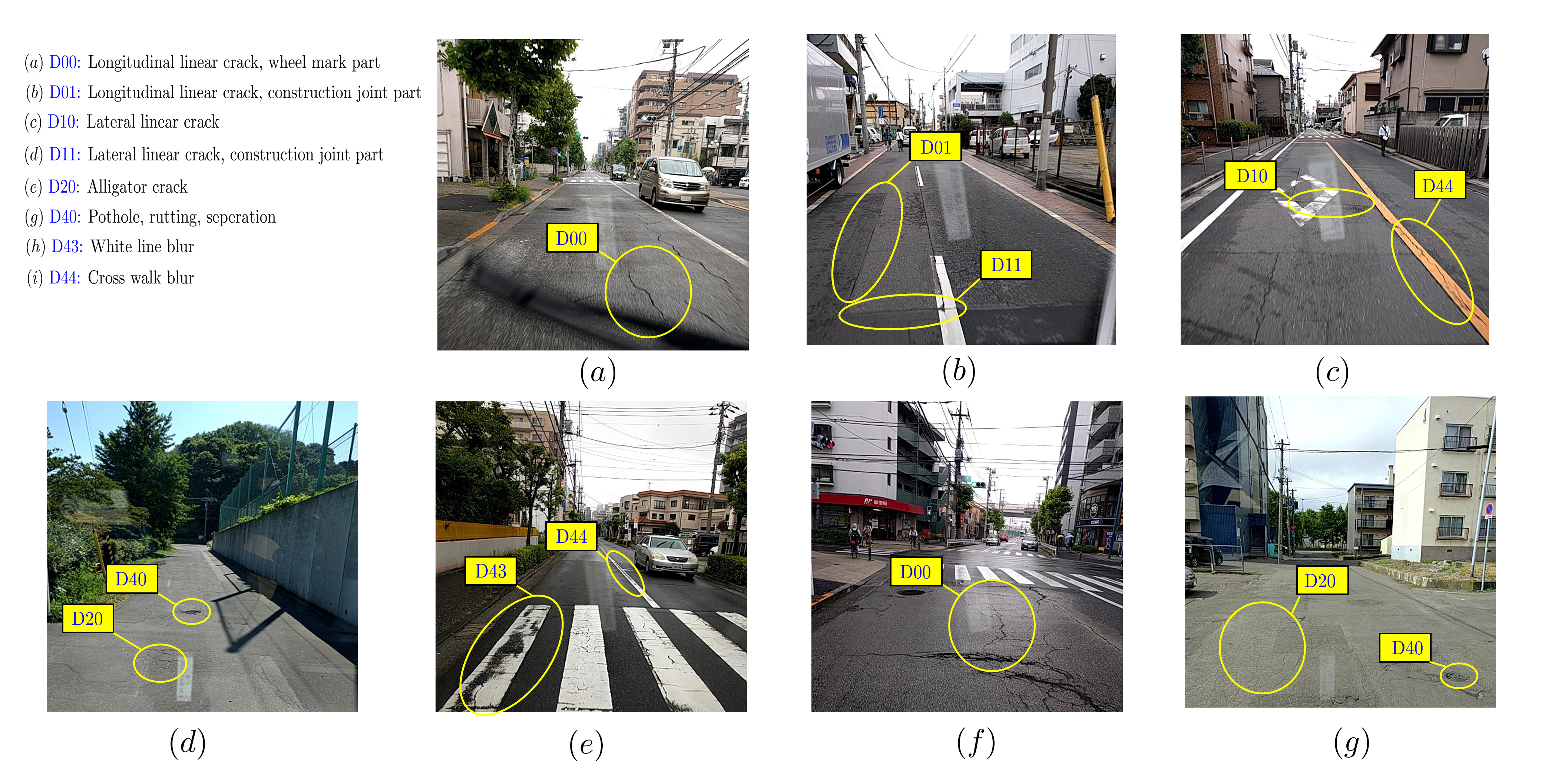}
	\caption{Sample images from RDD-2018 dataset \citep{maeda2018road}: (a) to (g) correspond to each of the eight categories with the legends.}
	\label{Fig-0}
\end{figure}
To avoid biases, the images have been photographed in various weather and illumination conditions from different regions of Japan including Chiba, Muroran, Ichihara, Sumida, Nagakute, and Numazu. 
During annotation, professional road damage expertise has been employed to verify various damage classes that ensure the reliability of the dataset. Various damage types and corresponding class identifiers have been listed in Table. \ref{T-1}. Each type of damages
have been  illustrated  in Fig. \ref{Fig-0}.
Primarily, the damage has been 
classified into cracks or different corruptions. Then, the
cracks have been divided into linear and alligator cracks. Whereas, other corruptions include both
potholes and rutting as well as other road damage classes such
as blurring of white lines.
\begin{table}
	\centering
	\caption{Various road damage types and corresponding class identifiers in RDD-2018 dataset \citep{maeda2018road}. }
	\begin{tabular}{c c c c }
		%\hline
		\\[-0.5em]
		\hline
		\\[-0.8em] % Adds extra space after hline
		\vtop{\hbox{\strut Class }\hbox{Identifier }}   &\vtop{\hbox{\strut Damage }\hbox{type }}  & \quad Alignment  & Details 
		\\[0.5em]
		\\
		\hline
		\\[-0.5em]
		D00 & Linear Crack & Longitudinal & Wheel-marked part 
		\\
		\\[-0.5em]
		D01 & Linear Crack & Longitudinal & Construction joint part 
		\\
		\\[-0.5em]
		\hline
		\\[-0.5em]
		D10 & Linear Crack & Lateral  & Equal interval 
		\\
		\\[-0.5em]
		D11 & Linear Crack & Lateral & Construction joint part 
		\\
		\\[-0.5em]
		\hline
		\\[-0.5em]
		D20 & Alligator Crack & - & Partial pavement, overall
		pavement
		\\
		\\[-0.5em]
		\hline
		\\[-0.5em]
		D40 & Other Crack & - & Pothole, rutting, separation
		\\
		\\[-0.5em]
		D43 & Other Crack & - &  White line blur
		\\
		\\[-0.5em]
		D44 & Other Crack & - & Cross walk blur
		\\
		\\[-0.5em]
		\hline
	\end{tabular}
	\label{T-1}
\end{table}
%	For the variability and challenges in the datasets, we have included images that
%	characterize limited and/or full illumination, low visibility,  high degree of occultation, multiple objects with overlap, complex backgrounds, textural similarity of the object and the background, and noisy environment.
%	Additionally, the  images of the
%	dataset have variations in their scale, orientation, and resolution. 
%The datasets can be available publicly in the following link: ?
%	\begin{figure}
	%			\centering
	%			\includegraphics[width=\textwidth]{Fig-1.pdf}
	%			\caption{(a)  Representative samples images  from  endangered wildlife dataset that consist of  eight  classes:  (a) Polar Bear; (b) Galápagos Penguin; 
		%					(c) Giant Panda; 
		%					(d) Red Panda; 	
		%					(e) African forest elephant; 
		%					(f) Sunda Tiger;
		%					(g) Black Rhino;  and (h)
		%					African wild Dog}
	%			\label{Fig-1}
	%		\end{figure}
%	\\
%	\\
%	\\
%	\\
\\	
\\
{\bf 3. Proposed  Methodology for damage detection:}
\\
\\	
In object detection, target object
classification and localization are performed simultaneously
where the target class has been categorized and separated from
the background. The purpose of object localization is to
locate objects by drawing bounding boxes (BBs) on input images containing the entire object. This is particularly useful for counting endangered species for accurate surveying.     
To this end, the main goal of the current work is to develop an efficient and robust damage classification and accurate damage localization model.
In this regard,  different variants of YOLO  \citep{Redmon_et_all-2016,Redmon_Farhadi-2017,Redmon_et_all-2018,Bochkovskiy_et_all-2020}  are some of the best high-precision one-stage object detection models.
More recently, YOLOv5 \citep{jocher2021ultralytics} has been introduced that
currently achieves the best  detection performance and  has four different model variants including YOLOv5s,
YOLOv5m, YOLOv5l, and YOLOv5x depend on different model widths and depths. 
In general, the overall architecture of YOLOv5 consists of the following parts: a backbone for deep feature extraction, followed by the neck for gathered semantic feature fusion, and finally head network for object classification and localization.
The original version of YOLOv5  utilizes CSPDarknet53 \citep{wang2020cspnet,Bochkovskiy_et_all-2020} with SPP and PANet \citep{Liu-IEEE-2018} as backbone and neck, respectively. Whereas, YOLO detection head \citep{Redmon_et_all-2016} has been employed in the detection head.
\begin{figure}
	\centering
	\includegraphics[width=\textwidth]{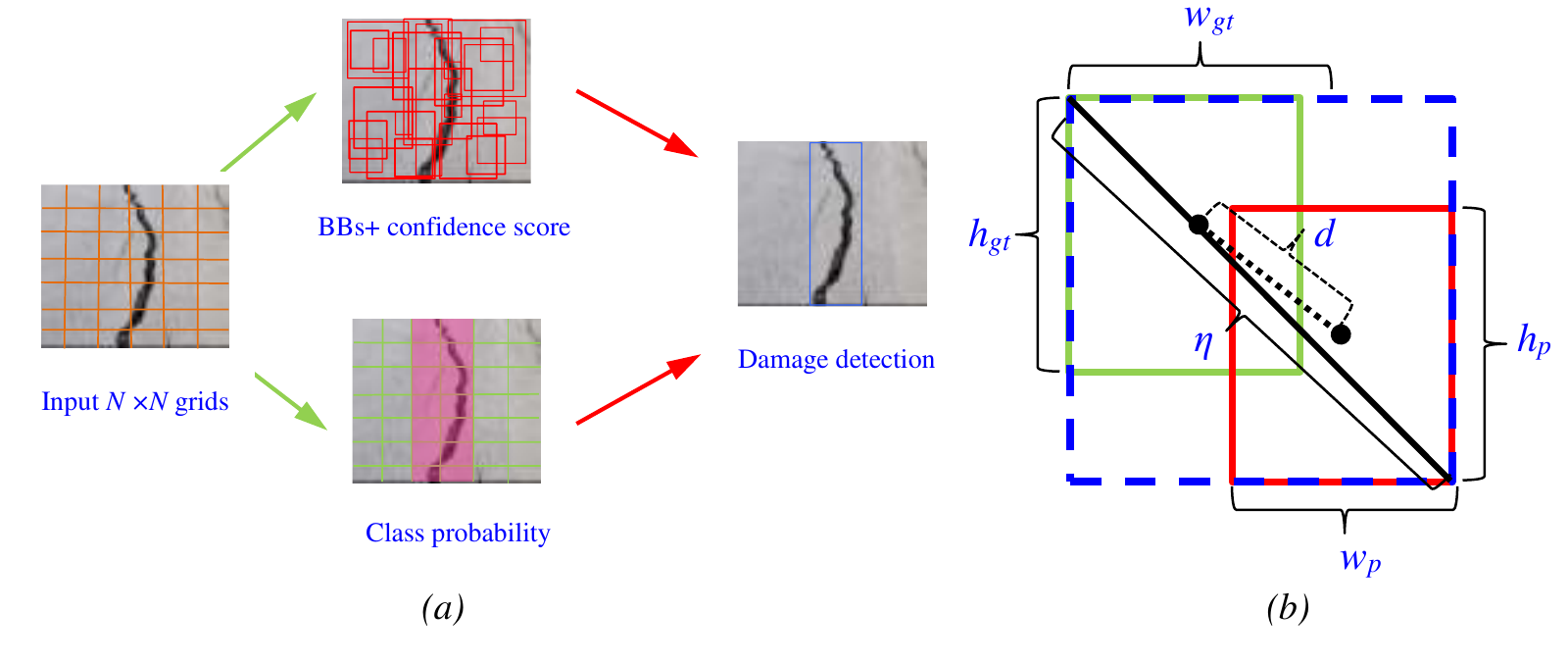}
	\caption{Schematic of (a) YOLO object localization process for damage localization; (b) Schematic of CIoU offset regression for target BBs predictions.}
	\label{Fig-1}
\end{figure}
The YOLO model transforms the object detection task into a regression problem by generating BBs coordinates and probabilities for each class as shown in Fig. \ref{Fig-1}. 
During the process, the inputted image size has been uniformly divided into $N \times N$  grids where $B$ predictive BBs have been generated. Subsequently, a confidence score has been assigned if the target object falls inside that particular grid. It detects the target object for a particular  class when 
the center of the ground truth lies inside a specified grid. 
During detection, each grid predicts $N_B$ numbers of  BBs with the confidence value $\Phi_B$   as: 
\begin{equation}
	\Phi_B=\mathcal{P}_r(obj)\times \mbox{IoU}^{t}_{p} \quad \quad  \, \vee \,\mathcal{P}_r(obj) \in {0, 1}
	\label{E-1}
\end{equation} 
where $\mathcal{P}_r(obj)$ infers the accuracy of BB prediction, i.e., $\mathcal{P}_r(obj)=1$ indicates that the  target class falls inside the
grid, otherwise, $\mathcal{P}_r(obj)=0$. 
The degree of overlap between ground truth  and the predicted
BB has been described by the scale-invariant evaluation metric intersection over union (IoU) which can be expressed as  
\begin{equation}
	\mbox{IoU}=\frac{\mathbf { B}_p\,\,\cap\,\,\mathbf {B}_{gt}}{\mathbf { B}_p\,\,\cup\,\,\mathbf {B}_{gt}}
	\label{E-2}
\end{equation}
where $\mathbf {B}_{gt}$ and $\mathbf {B}_{p}$ are the  ground truth and predicted BBs, respectively. 
\\
\\
{\bf  3.1 Loss in BBs regression :}
\\
\\
To further improve BBs regression and gradient disappearance, generalized IoU (GIoU) \citep{Rezatofighi_et_all-2019} and distance-IoU (DIoU) \citep{Zheng_et_all-2020} as been introduced considering aspect ratios and orientation of the overlapping BBs.
More recently, complete IoU (CIoU) \citep{Zheng_et_all-2020} has been proposed for improved accuracy  and faster convergence speed in BB prediction which  can be expressed as 
\begin{eqnarray}
	\label{E-3}
	&&\mathcal{L}_{\mbox{CIoU}}= 1+\beta \xi+ \frac{\alpha^2(\mathbf {b_p, b_{gt}})}{\eta^2}-\mbox{IoU}
	\\
	\label{E-4}
	&& \xi= \frac{4}{\pi^2}  \left(\tan^{-1}\,\frac{w_{gt}}{h_{gt}}-\tan^{-1}\,\frac{w_p}{h_p}\right)^2;\quad \beta= \frac{\xi}{(1-\mbox{IoU})+\xi^{'}}
\end{eqnarray}
where $\mathbf {b_{gt}}$ and  $\mathbf {b}_p$    denotes the  centroids  of $\mathbf {B}_{gt}$ and $\mathbf {B}_p$, respectively; $\xi$ and $\beta$  are  the consistency  and  trade-off parameters, respectively.
As shown in Fig. \ref{Fig-2} -(b), $\eta$ is the smallest diagonal  length of $\mathbf { B}_p\,\,\cup\,\,\mathbf {B}_{gt}$;  
$w_{gt}$, $w_p$ are widths and $h_{gt}$, $h_p$ are heights of $\mathbf {B}_{gt}$ and $\mathbf {B}_p$, respectively. 
With increasing  $w_p/h_p$, we get $\xi\rightarrow0$ from Eq. \ref{E-4}. 
Therefore, to optimize the influence of   $\xi$ on the CIoU, $w_p/h_p$ can be properly chosen for the detection model. 
Finally, the best BB prediction can be obtained from the non-maximum suppression (NMS) \citep{Ren_et_al-2015} algorithm from various scales. In object detection in the YOLO framework, 
The total loss function $\Delta_l$ consist of BB coordinate prediction error $\Delta_{cor}$,  IoU error $\Delta_{IoU}$, classification error term $\Delta_{cl}$.
The loss function $\Delta_l$ for BBs regression  can be formulated  as,
\begin{equation}
	\Delta_l= \Delta_{cor}+\Delta_{cl}+ \Delta_{IoU}.
	\label{E-5}
\end{equation} 
\\
\\
{\bf 3.2 Performance metrics:} 
\\
\\
In the present work, the performance of the object detection models has been evaluated by common standard measures \citep{ferri2009experimental} including average precision (AP),  precision (P),  recall (R), IoU, F-1 score, mean average precision (mAP), etc. 
The confusion matrix obtained from the evaluation procedure provides the following interpretations of the test results: true positive (TP), false positive (FP), false negative (FN), and true negative (TN).
During object classification, the classified object  can be defined
as TP for IoU $\ge 0.5$. Whereas, it can be classified as FP for IoU $<0.5$. 
Based on the aforementioned  interpretations, 
the metric P  of the classifier can be  defined by its  ability to distinguish target classes correctly as :
\begin{equation}
	\mbox{P}= \frac{\mbox{TP}}{(\mbox{TP+FP})};
	\label{E6}
\end{equation}
The ratio of the correct prediction of target classes is called R of the classifier which can be  evaluated  as:
\begin{equation}
	\mbox{R}=\frac{\mbox{TP}}{(\mbox{TP+FN})}
	\label{E7}
\end{equation}
The higher values of P and R indicate superior detection capability. 
Whereas, F-1 score is the arithmetic mean of the P and R  given as :
\begin{equation}
	\mbox{F1} = 2 \times \bigg(\frac{\mbox{P R}}{\mbox{P + R}}\bigg)
	\label{E8}
\end{equation}
A relatively high  F1 score represents a robust detection model.
The performance metrics AP can be defined as the area under a P-R curve \citep{davis2006relationship}  as follows 
\begin{equation}
	\mbox{AP}= \int_0^1 \! \mbox{P (R)} \, \mathrm{d}\mbox{R}
	\label{E9}
\end{equation}
A higher average AP value indicates better accuracy in predicting various object classes. In addition, 
AP$_{50:95}$ denotes  AP over IoU=$0.50:0.05:0.95$; AP$_{50}$ and AP$_{75}$ are APs  at IoU threshold of $50\%$ and $75\% $, respectively.  
The AP for detecting small, medium, and large objects can be measured through 
AP$_{S}$, AP$_{M}$, and AP$_{L}$, respectively. 
Finally, mAP  can be obtained from  the average of all APs as: 
\begin{equation}
	\mbox{mAP}= \frac{1}{N_c} \sum_{i=1}^N  	\mbox{AP}_i.
	\label{E10}
\end{equation} 
In YOLOv5, various combinations of activation functions including sigmoid, leaky-
ReLU and SiLU \citep{hendrycks2016gaussian} can be utilized to improve the performance of the model for a specific detection task. In addition, bag of freebies and specials \citep{Bochkovskiy_et_all-2020} can also be employed to further optimize the detection architecture of YOLOv5. As the latest generation of
the YOLO series, YOLOv5 has been shown to provide state-of-the-art detection performance. Therefore, it has been chosen as the baseline model for our present study. 
\\
\\
{\bf 3.3  DenseSPH-YOLOv5 architecture:}
\\
\\	
In recent years, various attempts have been made on  DL-based computer vision models for damage detection such as  	Faster R-CNN 	\citep{kluger2018region,wang2018road},  SSD \citep{maeda2018road,wang2018deep}, RetinaNet \citep{angulo2019road}, YOLO \citep{alfarrarjeh2018deep, mandal2020deep}, YOLOv2 \citep{majidifard2020pavement}, YOLOv5 \citep{guo2022road} etc. 
Although the aforementioned techniques have demonstrated outstanding performance, however, the damage  detection task  faces several  challenges, in particular,
due to the presence of complex and noisy backgrounds, significant variability of lightening conditions, low visibility, densely packed classes, and overlap, the coexistence of multi-object classes with various aspect ratios, and other morphological characteristics \citep{azimi2020data,naddaf2020efficient}. 
In this regard,  YOLOv5 can be a suitable model which demonstrates both superior accuracy and faster detection speed compared to YOLOv4. The state-of-the-art YOLOv5 is comparatively light, yet effective, which performs multiple downsampling and turns shallow semantic information
into high-level semantic information that effectively reduces the number of parameters. However, this inevitably loses
semantic information and makes the network ineffective at
extracting and fusing discriminating semantic features. Therefore, the original YOLOv5 network requires further improvement due to its insufficient fine-grain contextual feature extraction capability leading to missed detection and false object predictions for various damages/cracks which possess a wide range of textures, shapes, sizes, and colors \citep{cao2020survey,azimi2020data,naddaf2020efficient}. 
In order to  achieve
high accuracy and high efficiency in damage detection,  we propose a novel object localization algorithm DenseSPH-YOLOv5 based on a state-of-the-art  YOLOv5 network to enhance feature extraction, preserve fine-grain localized information and improve feature fusion that provides superior damage detection under various challenging environments. 
The overall network of the DenseSPH-YOLOv5 model is shown in  Fig. \ref{Fig-3}. 
To improve performance in terms of classification accuracy and object localization, we perform extensive experiments, and various modifications are proposed which are detailed in the subsequent sections.
\\
\\
{\bf 3.3.1 DenseNet block :}
Since YOLOv5 reduces the feature maps from the inputted images through convolution and down-sampling procedures that result in significant semantic feature loss during transmission. 
To this end, we have introduced DenseNet \citep{Huang-IEEE-2017} in the original CSPDarknet53 of YOLOv5 attached to the CSP module to preserve critical feature maps and efficiently reuse the discriminative feature information as shown in Fig. \ref{Fig-3}.
In DenseNet, each layer has been connected to other layers in a feed-forward mode where $n$-th layer can  receive the  important  feature information $\xi_n$  from all the previous layers $\xi_0, \xi_1,..., \xi_{n-1}$ as:
\begin{equation}
	\xi_n= \Theta_n[\xi_0,\, \xi_1,\,...,\xi_{n-1}]
	\label{E-11}
\end{equation}
where  $\Theta_n$ is the feature map function for $n$-th layer.  The schematic of the DenseNet blocks network structure
have been shown in Fig. \ref{Fig-4}-(d).
Through our extensive experiments, we found out that DenseNet improves the feature transfer and mitigates over-fitting in the detection network. 
Therefore, in the proposed Dense-SPH YOLOv5 network, as shown in Fig. \ref{Fig-2},   we have introduced four DenseNet blocks: the first block (Dense B-1) has been attached before cross-stage partial block CSP3;  the second  block (Dense B-2)
has been placed before CSP6;  whereas, third (Dense B-3) and fourth (Dense B-4) blocks have been added before CSP6 and CSP3, respectively which results in enhanced feature propagation.
Additionally, by reducing redundant feature operations, the Dense-CSPDarknet53 network improves the computational speed.
\\
\\
{\bf 3.3.2  Convolutional block attention module (CBAM):}  
Convolutional block attention module (CBAM) \citep{woo2018cbam}  is a lightweight, yet effective attention block which can be integrated into the object detection model in an end-to-end manner that has shown superior detection accuracy \citep{zhu2021tph}.   For an inputted feature map, CBAM sequentially employs the attention map along the channel and spatial dimensions.	
As shown in Fig. \ref{Fig-4}-(a), the network structure of the CBAM module consists of two sequential sub-modules:   Channel Attention Module (CAM) and  Spatial  Attention Module (SAM) for adaptive refinement of the feature map at every intermediate convolution blocks by multiplying input feature map with the attention map. 
For inputted  feature map $\fg {F_c} \in  \mathbb{R}^{C \times H \times W}$, CAM produces 
1D channel attention map $\fg {M^c} \in  \mathbb{R}^{C \times 1 \times 1}$, and sequentially, SAM infers 2D spatial attention
map $\fg {M^s} \in  \mathbb{R}^{1 \times H \times W}$ as shown in Figs. \ref{Fig-3}-(a-c). 
The overall CBAM attention process can
be expressed  as:
\begin{eqnarray}
	\fg {F_c^{'}}= \fg {M^c} (\fg {F_c}) \, \tenp\, \fg {F_c}; \quad \fg {F_c^{''}}= \fg {M^s} (\fg {F_c^{'}}) \, \tenp\, \fg {F_c^{'}};
	\label{E-12}
\end{eqnarray}
where ; $\fg {F_c^{''}}$ is the CBAM  refined
output; $\tenp$ represents  element-wise multiplication. More specifically, CAM produces two different spatial descriptors  including average-pooled features  $\fg {F^c_{avg}}$ and max pooled
features $\fg {F^c_{max}}$ which produce  channel attention map $\fg {M^c} \in  \mathbb{R}^{C \times 1 \times 1}$ from multi-layer perceptron (MLP). The CAM attention can be expressed as: 
\begin{eqnarray}
	\fg {M^c} (\fg {F_c})= \sigma (\fg {W_1} ( \fg {W_0}(\fg {F^c_{avg}})))
	+ \fg {W_1} ( \fg {W_0}(\fg {F^c_{max}}))
	\label{E-13}
\end{eqnarray}
where $\fg {W_0} \in  \mathbb{R}^{C/r \times C}$, and $\fg {W_1} \in  \mathbb{R}^{C \times C/r}$ are the MLP weights ; $\sigma$ represents  the sigmoid activation  function. In SAM, inter-spatial  attention map $\fg {M^s} (\fg {F_c})\in  \mathbb{R}^{H \times W}$ has been aggregated to generate  2D maps: $\fg {F^s_{avg}}\in  \mathbb{R}^{1 \times H \times W}$ and $\fg {F^s_{max}}\in  \mathbb{R}^{1 \times H \times W}$. From these two feature maps, the spatial attention can be computed as: 
\begin{eqnarray}
	\fg {M_s} (\fg {F})= \sigma ( f^{n \times n} ([\fg {F^s_{avg}} ; \fg {F^s_{max}}])
	\label{E-14}
\end{eqnarray}
where  $f^{n \times n}$ denotes  a convolution operation with the filter size of ${n \times n}$ with $n=7$.
The implementation of the CBAM module into the proposed DenseSPH-YOLOv5  significantly improves the detection accuracy of the damage detection by providing specific attention to the dense objects and confusing noisy areas which proved the effectiveness of this module. Overall, it helps to learn more expressive features that demonstrate significant improvement in detection accuracy for road damage datasets considered herein. 
\\
\\
{\bf 3.3.3 Swin transformer  encoder blocks: } 
Inspired by the superior performance of the vision transformer
\citep{dosovitskiy2020image} in dense and occluded object detection, in the proposed model, Swin
transformer (STR) encoder blocks  \citep{liu2021swin} have been fused to all four detection heads of DenseSPH-YOLOv5  architecture as shown in Fig. \ref{Fig-2}.  Such implementation improves the global semantic feature extraction and contextual information fusion leveraging self-attention mechanism \citep{vaswani2017attention}  that demonstrated superior performance in dense object detection \citep{gong2022swin}. 

\begin{figure}
	\centering
	\includegraphics[width=1.1\textwidth]{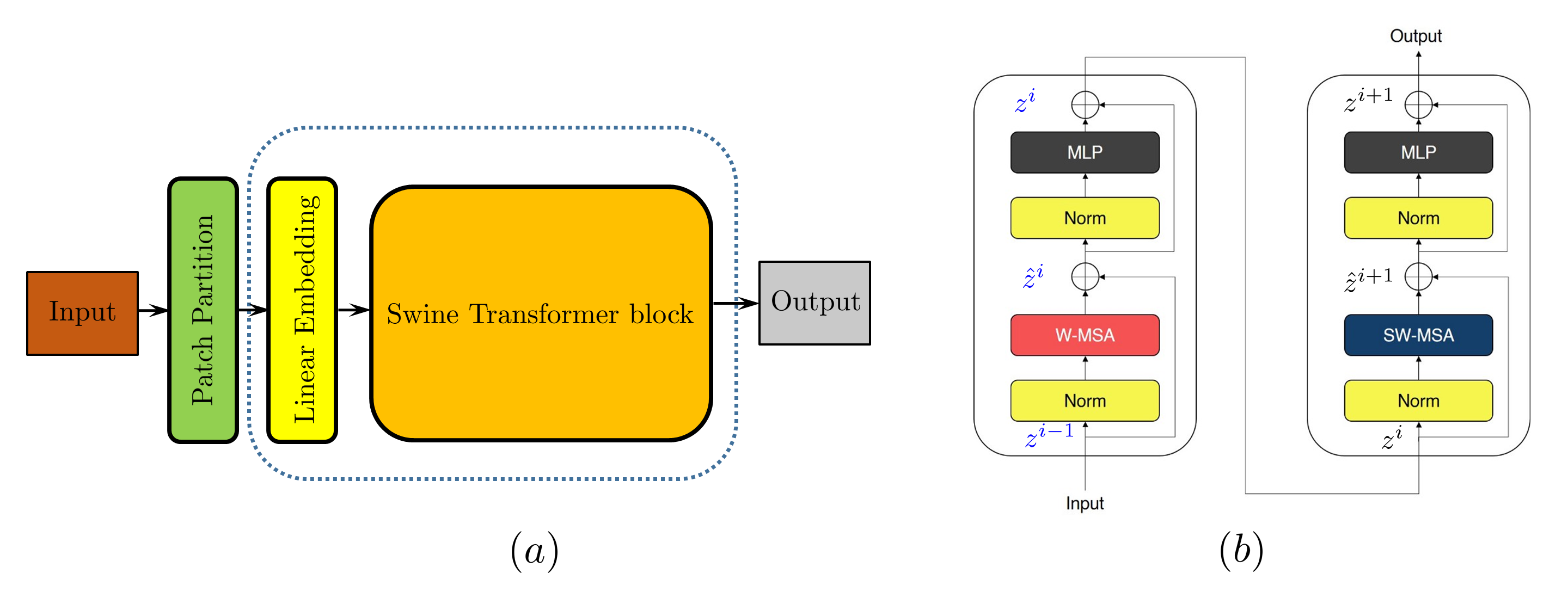}
	\caption{Schematic of   (a) STR encoder architecture that contains (b) regular W-MSA and SW-MSA used in detection heads of  DenseSPH-YOLOv5 architecture. }
	\label{Fig-5}
\end{figure}

\begin{figure}
\noindent
\centering
\includegraphics[width=1.1\linewidth]{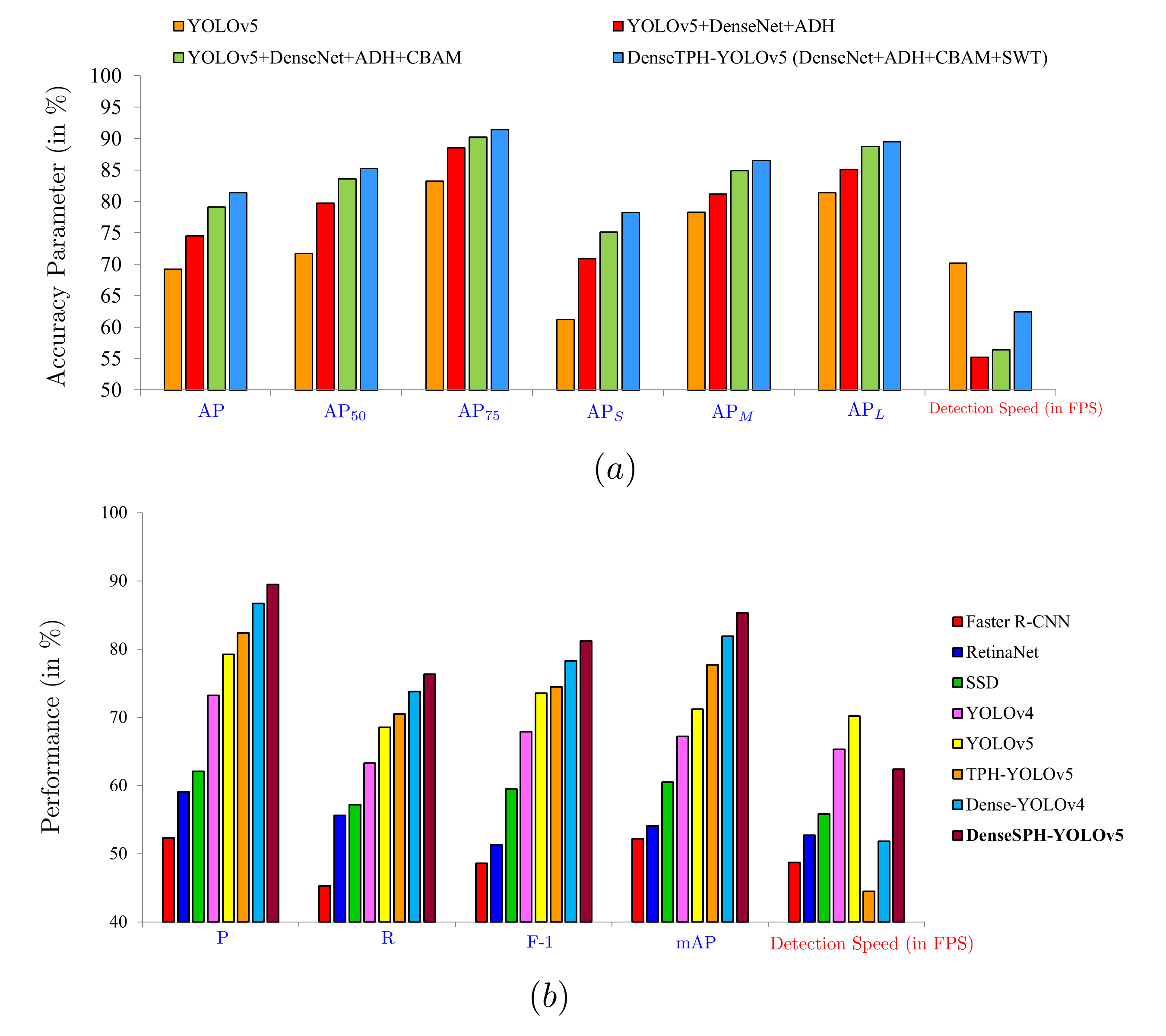}
\caption{\label{Fig-6} Comparison bar chart of precision parameters and detection speed for  (a) different combinations of backbone and neck of the detection model;  (b) different state-of-the-art models and DenseSPH-YOLOv5. }
\end{figure}

For inputted feature map   $\fg {F_s} \in  \mathbb{R}^{H \times W \times C}$, it transforms to $\fg {Q^s, K^s, V^s} \in  \mathbb{R}^{N \times  C'}$ where $N = H \times W$ to feed Multi-head Self Attention (MSA) module after linear projection and reshape operations. The output feature map $\fg {Z^s}$ from MSA aggregates global information which  can be expressed as 
\begin{eqnarray}
	\fg {Z^s} = \fg {A^s V^s}; \quad \fg {A^s} = \mbox{softmax} \,\, \fg {Q^s {K^s}^T};
	\label{E-15}
\end{eqnarray}
where $\fg {A^s} \in  \mathbb{R}^{N \times  N}$ is the self-attention matrix that represents the relationship between feature map elements with remaining elements.  Although, MSA in Transformer is effective in generating a self-attention matrix by integrating multiple independent subspaces, however, the global computation overhead is significantly high for high-resolution images. Thus, for dense prediction or to tackle high-resolution images, the computational complexity of MSA in Transformer is not suitable which leads to
quadratic complexity with respect to the number of tokens \citep{liu2021swin}.  To this end, STR can significantly improve the computational efficiency of MSA that has linear computational complexity
with respect to image size which enhances the performance of the model in terms of detection speed and accuracy. Therefore, in the proposed DenseSPH-YOLOv5 detection model, the STR encoder module has been fused into four different prediction heads. As shown in Fig. \ref{Fig-5} -(b), each  STR  encoder contains two sub-layers that include shifted window-based multi-head self-attention (MSA) module,  followed by a fully-connected MLP with GeLU nonlinearity. Residual	connections are used after each MSA module. Subsequently,  Layer Norm (LN)  has been added before  MSA and  MLP. 
In STR, the obtained feature map has been dived into non-overlapping separate windows in the W-MSA module. Subsequently, self-attention has been computed in each local window. Applying SW-MSA partitioning, the consecutive
STR blocks can be  computed as: 
\bey
\fg{\hat{z}}^i&=&\mbox{\bf{W-MSA}} (\mbox{\bf{LN}}(\fg{{z}}^{i-1}))+ \fg{{z}}^{i-1}; \quad
\fg{{z}}^i=\mbox{\bf{MLP}} (\mbox{\bf{LN}}(\fg{\hat{z}}^{i}))+ \fg{{\hat{z}}}^{i}
\label{E-16}\\
\fg{\hat{z}}^{i+1}&=&\mbox{\bf{SW-MSA}} (\mbox{\bf{LN}}(\fg{{z}}^{i}))+ \fg{{z}}^{i}; \quad 
\fg{{z}}^{i+1}=\mbox{\bf{MLP}} (\mbox{\bf{LN}}(\fg{\hat{z}}^{i+1}))+ \fg{{\hat{z}}}^{i+1}
\label{E-17}
\eey
where $\fg{{z}}^{i}$ and $\fg{\hat{z}}^{i}$ represent the output features from MLP and SW-MSA modules, respectively.  
For a feature map $\fg {F_s} \in  \mathbb{R}^{H \times W \times C}$ with $m \times m$  size of local window, the computational complexity $\Upsilon$ can be obtained as 
\begin{eqnarray}
	\Upsilon (\mbox{\bf{MSA}}) = 4HWC^2 + 2(HW)^2C; \quad \Upsilon (\mbox{\bf{W-MSA}}) = 4HWC^2 + 2(HW)M^2C
	\label{E-18}
\end{eqnarray}
where $HW$ is the patch number. 
Evidently, for large  $HW$, the computational overhead of global MSA is exceptionally high, while, W-MSA is scalable. Thus, implementing  STR significantly reduces the computational complexity of the model.  
%	{\bf 4.2  Improvement of discriminative feature extraction:}
%	\\
%	In the present study, we have introduced a new residual block CSPX1-$n$ where $n$ represents residual weighting operations to improve detection speed and performance.    
%	We integrate  CSPX1-$n$ modules in the CSPDarknet53 backbone replacing the original CSP8 and CSP4 residual blocks to extract fine-grained rich semantic information as shown in  Fig. \ref{Fig-3}. 
%	In the CSPX1-$n$ block, we divide the input features into two parts.
%	In the first part, $(3\times3)$ convolution was performed followed by an additional  $(3\times3)$ convolution to maintain the number of feature maps after entering the next residual unit as shown in  Fig. \ref{Fig-4}(c). 
%	To further improve the feature extraction, we perform $3\times3$ convolution at the end. 
%	Whereas, the second part acts as a residual edge for the convolution. 
%	These two parts have been concatenated at the end to improve the semantic feature information. 
%	Implementation of the CSPX1-$n$ modules in the improved  CSPDarknet53 helps to learn more expressive features that demonstrate significant improvement of detection accuracy for the custom wildlife datasets used herein. 
%As the last three layers contain relatively higher semantic information, these are passed to the SPP and the PANet. 
%The last feature layers contain the finest feature information, and is connected to the SPP. 
%The other two layers are integrated into the PANet as shown in Fig. \ref{Fig-3}. 
\\
\\
{\bf 3.3.4 Receptive field enhancement:}
One of the requirements of CNN  is to have fixed-size input images. However, due to the different aspect ratios of the images, they have been fixed by cropping and warping during the convolution process which results in losing important features.  
In this regard,   SPP \citep{He-IEEE-2015}  applies an efficient strategy in detecting target objects at multiple length scales. 
To this end,  we have added an SPP block integrated with DCSP-3 in the Dense-CSPDarknet53 backbone to improve 
receptive field representation and extraction of important contextual features as shown in Fig. \ref{Fig-3}. 	 
\\
\\	
{\bf 4. Results:}
\\
\\
In this section, the performance and detection accuracy of the proposed DenseSPH-YOLOv5 framework have been discussed that been evaluated in the  RDD-2018  dataset consisting of 8 different categories of damage classes. For better clarity in BB representation, each damage type has been marked with the corresponding class identifiers as shown in Table \ref{T-1}. 
The performance of the DenseSPH-YOLOv5 network has been optimized through extensive ablation studies.  Finally, the performance of the proposed model has been studied in detail and compared with several state-of-the-art object detection models. 
\\
\\
{\bf 4.1 Training procedure :} 
\\
\\
In the present work, we have performed an extensive and elaborate study to explore the comparative performance analysis of the proposed DenseSPH-YOLOv5 models for road damage localization tasks.
From the RDD-2018 dataset, a total of $80 \%$  and $20 \%$ images have been randomly
chosen for training and validation, respectively. 
%	Thus, the training
%	data included 7,240 images, and the evaluation data had
%	1,813 images.
For all the experiments, we have used a Windows 10 Pro (64-bit) based computational system that has  Intel Core i5-10210U    with  CPU $@$ 2.8 GHz $\times 6$,  32 GB DDR4 memory, NVIDIA GeForce RTX 2080  utilizing  GPU parallelization. 
%As required CV libraries, Visual Studio v15.9 (2017), and OpenCV 4.5.1-vc14 have been integrated with DarkNet. 	
As part of data augmentation, along with traditional methods such as photometric distortion and geometric distortions,  
additional data augmentation strategies including MixUp \citep{zhang2017mixup}, CutMix \citep{yun2019cutmix} and Mosaic \citep{Bochkovskiy_et_all-2020} have been combined which help improves the performance of DenseSPH-YOLOv5. In addition, Mosaic augmentation can significantly enrich the background information. Since such a method increases the batch size, therefor,  batch normalization has been followed.
Unless otherwise stated, a  batch size set to 16  with a total number of training steps has been kept as  80.  The initial learning rate has been set to 0.01 with SGD
optimizer. The training dataset has been trained to utilize the available pre-trained weights-file \citep{lin2014microsoft}.  
\\
\\
{\bf 4.2 Optimization of network performance:}
\\
\\
At first, we conduct extensive experiments to select proper backbone-neck combination modules to optimize the performance of the proposed DenseSPH-YOLOv5 model in terms of both detection accuracy and speed. 
For different combinations of backbone-neck configurations, detection accuracy in terms of parameters AP, AP$_{50}$, AP$_{75}$, AP$_{S}$, AP$_{M}$, and AP$_{L}$ as well as detection speed (in FPS) has been reported in Table. \ref{T-2}.
%For the comparison, we select ? as the activation function. 
From the Table. \ref{T-2}, one can see the introducing 
Additional Detection Head (ADH) improves performance by employing an additional feature fusion mechanism. 
Notably,  there are  2.3$\%$ increases in  AP$_{50}$; and  6.1$\%$ increases in  AP$_{L}$ compared to the baseline original YOLOv5. 
Furthermore, implementation of DenseNet with ADH enhances the performance further, in particular, for relatively small object detection with  3.6 $\%$ increase in  AP$_{L}$ as shown in Fig. \ref{Fig-6}-(a).  However, this results in a signification reduction in detection speed (from 69.1 FPS to 55.2 FPS).
With the introduction of the CBAM in the neck part of the detection model, the detection performance improves further compared to DensNet+ADH configuration for relatively medium and large object sizes with 3.7$\%$ and 3.6$\%$ increase in AP$_{M}$ and AP$_{L}$, respectively while slightly compromising the detection speed. 
\begin{table}
	\centering
	\caption{Performance of various residual and dense block combinations in DenseSPH-YOLOv5 architecture for anchors size of  $416\times 416$.
	}
	\begin{tabular}{c c c c c c c c c c }
		%\hline
		\\[-0.5em]
		\hline
		\\[-0.8em] % Adds extra space after hline
		\vtop{\hbox{\strut Backbone}\hbox{\strut add-in}}& \vtop{\hbox{\strut Neck}\hbox{\strut add-in}}  & $AP$  & $AP_{50}$ & AP$_{75}$ & AP$_{S}$ & AP$_{M}$ & AP$_{L}$ & FPS
		\\[-0.0em]
		\hline
		\\[-0.5em]
		(YOLOv5) \quad- & - & 69.2 & 71.7 & 83.2  & 61.2  & 78.3  & 81.4 & 70.2 
		\\
		\\[-0.5em]
		- & ADH  & 70.4 & 74.1 & 85.9  & 67.3  & 79.7  & 84.7 & 69.1
		\\
		\\[-0.5em]
		DenseNet & ADH & 74.5 & 79.7 & 88.5  & 70.9  & 81.2  & 85.1 & 55.2
		\\
		\\[-0.5em]
		DenseNet & ADH+CBAM & 79.1 & 83.6 & 90.2  & 75.1 & 84.9  & 88.7 & 56.4
		\\
		\\[-0.5em]
		DenseNet & ADH+CBAM+STR & \bf{81.4}  & \bf{85.2} & \bf{91.4}  & \bf{78.2} & \bf{86.5}  & \bf{89.5} & \bf{62.4} 
		\\
		\\[-0.5em]
		\hline
		%\hline
	\end{tabular}
	%}
\label{T-2}
\end{table}
Thus, the implementation of CBAM has been proven to be 
an effective strategy to retain semantic and delicate spatial information from spatial and channel attention mechanisms that enhance the overall performance of the model.  
Finally,  the best performance has been achieved when both CBAM and STR respectively have been integrated into the neck and detection head which demonstrate significant improvements in the accuracy parameter, in particular,  AP, AP$_{50}$, AP$_{75}$, AP$_{M}$ and AP$_{L}$ increase by 12.2$\%$, 13.5$\%$, 8.2$\%$, 8.3$\%$,  and 8.1$\%$, respectively compared to original YOLOv5 configuration. 
Moreover, we have also observed improvement in detection speed with the introduction of STR in the detection head by reducing the computational complexity compared to only ADH+CBAM configuration in the neck as shown in Fig. \ref{Fig-6}-(a).
Therefore, a such  configuration  in DenseSPH-YOLOv5
provides the optimal performance in terms of detection accuracy
and speed for the road-damaged data set considered herein.
In summary, together with proper activation function and improved backbone-neck combination in DenseSPH-YOLOv5 provides an efficient high-performance model for damage detection in complex scenarios. 
\begin{table}
\centering
\caption{Comparison of different performance parameters including  P, R, F1, mAP, and detection speed (in FPS) between  DenseSPH-YOLOv5 and other SOAT models where bold highlights the best performance values.}
\begin{tabular}{c c c c c c c c }
	%\hline
	\\[-0.5em]
	\hline
	\\[-0.8em] % Adds extra space after hline
	Model & P ($\%$) & R ($\%$)  &F1-score ($\%$)  & mAP ($\%$)  & Dect. time (ms) & FPS
	\\[-0.0em]
	\hline
	\\[-0.5em]
	Faster R-CNN & 52.32 & 45.39 & 48.60  & 52.17  & 20.50 & 48.7  
	\\
	\\[-0.5em]
	RetinaNet & 59.11 & 55.67 & 51.35  & 54.11  & 18.87 & 52.7 
	\\
	\\[-0.5em]
	SSD & 62.13 & 57.19  & 59.56  & 60.52  & 17.92 & 55.8
	\\
	\\[-0.5em]
	YOLOv4 & 73.22 & 63.35 & 67.93  & 67.13 & 15.31 & 65.3 
	\\
	\\[-0.5em]
	YOLOv5 & 79.17 & 68.47 & 73.43  & 71.13 & \bf{14.24} &  \bf{70.2 } 
	\\
	\\[-0.5em]
	TPH-YOLOv5 & 82.37 & 70.51 & 74.52 & 77.62 & 22.47 & 44.5
	\\
	\\[-0.5em]
	Dense-YOLOv4	 & 86.75 & 73.75 & 78.28  & 81.87 &   19.30 & 51.8
	\\
	\\[-0.5em]
	\bf{DenseSPH-YOLOv5} & \bf{89.51} & \bf{76.25} & \bf{81.18}  & \bf{85.25} &16.02 & 62.4 
	\\
	\\[-0.5em]
	\hline
	%\hline
\end{tabular}
%}
\label{T-3}
\end{table}
\\
\\
{\bf 4.3 State-of-the-art comparison:}
\\
\\
In this section, the detection performance of the proposed DenseSPH-YOLOv5 has been compared with some of the existing state-of-the-art detection models \citep{Zhao-IEEE-2019}.
For the  performance comparison, we have considered  Faster R-CNN \citep{Ren_et_al-2015},  RetinaNet \citep{Lin1-IEEE-2017}, 
SSD \citep{Liu-ECCV-2016}, YOLOv4 \citep{Bochkovskiy_et_all-2020}, Dense-YOLOv4 \citep{roy2022real}, YOLOv5 \citep{jocher2021ultralytics}, and TPH-YOLOv5 \citep{zhu2021tph}  that are trained in RDD-2018 datset.
% utilizing OpenMMLab object detection toolbox \cite{MMDetection}. 
Comparison of different performance parameters including  P, R, F1-score, mAP, and detection speed obtained from these models have been presented in Table \ref{T-3}. 
The comparison reveals that the accuracy values obtained from Faster R-CNN, RetinaNet, and  SSD are quite inferior compared to YOLO variants as visually illustrated in the bar-chart plot in Fig. \ref{Fig-6}-(b). 
Between YOLOv4 and YOLOv5, YOLOv5 demonstrated better performance with a $5.95\%$  increase in P and  $4.01\%$ increase in mAP, respectively. 
We observe that the performance of Dense-YOLOv4 is superior to the TPH-YOLOv5 with  $4.38\%$, $3.24\%$,  $3.76\%$, and  $4.25\%$ increase in P, R, F1, and mAP, respectively. 
However, the proposed DenseSPH-YOLOv5 yields the best performance reaching the values of  $89.51\%$, $76.25\%$, $81.18\%$, and  $85.25\%$ in P, R, F1, and mAP, respectively as shown in Fig.\ref{Fig-6}- (b). 
Moreover, DenseSPH-YOLOv5 provides a superior real-time detection speed of  62.4 FPS  which is $28.68\%$ and $16.98\%$   higher than TPH-YOLOv5 and Dense-YOLOv4  models, respectively. 
In summary, DenseSPH-YOLOv5 outshines some of the best detection models in terms of both detection accuracy and speed illustrating its suitability for automated high-performance damage detection models.
\\
\\
{\bf 4.4 Comparison with existing state-of-the-art damage detection models:}
\\
\\
In addition, the performance of the DenseSPH-YOLOv5 has been compared with several existing state-of-the-art road damage detection models evaluated in RDD-2018 datasets. 
As shown in Table \ref{T-4}, we compared the performance from various DL models including SSD Inception v2 \citep{maeda2018road},  SSD MobileNet \citep{maeda2018road}, YOLO \citep{alfarrarjeh2018deep}, Faster R-CNN 	\citep{kluger2018region},  Faster R-CNN with ResNet-152 \citep{wang2018road}, Ensemble models with Faster R-CNN and SSD \citep{wang2018deep}, and RetinaNet \citep{angulo2019road} with our Dense-YOLOv4 and DenseSPH-YOLOv5 models. 
Note, in  \citep{angulo2019road} additional images were included in the RDD-2018 dataset to improve the detection performance. 
From the direct comparison, two current state-of-the-art models Faster R-CNN with ResNet-152 \citep{wang2018road} and  Ensemble models  \citep{wang2018deep} have reached the F1 value of  62.55\% that is 2.63\%  improvement over  SSD MobileNet \citep{maeda2018road}. While RetinaNet \citep{angulo2019road} illustrated significant performance improvement,  Dense-YOLOv4 performs better with F1 of 78.28\% in the original RDD-2018 dataset. 
Relative to the aforementioned models, our proposed model has achieved the best F1 value of  81.18 \% among current state-of-the-art damage detection models.
In terms of other precision metrics, there are 3.64\% and  10.51\% improvements in P and R compared to RetinaNet \citep{angulo2019road}, respectively.
Comparing detection speed, DenseSPH-YOLOv5 provides competitive performance compared to SSD Inception v2 \citep{maeda2018road} elucidates its superiority in real-time damage detection. 
\begin{table}
\centering
\caption{Comparison of different performance parameters between  DenseSPH-YOLOv5 and other SOAT road damage detection models evaluated in RDD-2018 dataset where bold highlights the best performance values.}
\begin{tabular}{c c c c }	
%\hline
\\[-0.5em]
\hline
\\[-0.8em] % Adds extra space after hline 
Reference & \vtop{\hbox{\strut Method}\hbox{\strut}}  & Performance & \vtop{\hbox{\strut Det. Speed}\hbox{\strut (FPS)}}   \\ \hline
\\
\cite{maeda2018road} & SSD Inception v2 & F-1: 52.61; P: 81.10; R: 38.97
& \bf{63.1}
\\
\\[-0.5em]
\cite{maeda2018road} & SSD MobileNet & F-1: 59.92; P: 81.11; R: 47.52
& 30.6
\\
\\[-0.5em]
\cite{alfarrarjeh2018deep} & YOLO & F-1: 62.00; P: -; R: -
& -
\\
\\[-0.5em]
\cite{kluger2018region} & Faster R-CNN & F-1: 61.00; P: -; R: -
& -
\\
\\[-0.5em]
\cite{wang2018road} &  \vtop{\hbox{\strut Faster R-CNN}\hbox{\strut (ResNet-152)}}
& F-1: 62.55; P: -; R: -
& -
\\
\\[-0.5em]
\cite{wang2018deep} & \vtop{\hbox{\strut \quad \quad \quad Ensemble}\hbox{\strut (Faster R-CNN+ SSD)}}
& F-1: 62.55; P: -; R: -
& -
\\
\\[-0.5em]
\cite{angulo2019road}$^*$ & \vtop{\hbox{\strut RetinaNet}\hbox{\strut}}
& F-1: 74.97; P: 85.87; R: 65.75
& -
\\
\\[-0.5em]
\citep{roy2022real} & \vtop{\hbox{\strut Dense-YOLOv4}\hbox{\strut}}
& F-1: 78.28; P: 86.75; R: 73.75
& 51.8
\\
\\[-0.5em]
Ours (current study) & \vtop{\hbox{\strut \bf{DenseSPH-YOLOv5}}\hbox{\strut}}
& \bf{F-1: 81.18; P: 89.51; R: 76.25}
& 62.4
\\
\\[-0.5em]
\hline
%\hline
\end{tabular}
\label{T-4}
\end{table}
\\
\\
{\bf 4.5  Overall performance of DenseSPH-YOLOv5:}
\\
\\ 
From section 4.3, we observed that TPH-YOLOv5, Dense-YOLOv4, and DenseSPH-YOLOv5 provide better performance compared to other SOAT models. Therefore,  these three models are closely compared in terms of mAP, F1, IoU,   final loss, and average detection time as shown in Table \ref{T-5}.
The proposed DenseSPH-YOLOv5 has achieved the highest average IoU value of 0.803 indicating superior BB accuracy during target detection compared to the other two models. 
Similarly, it has also illustrated better detection performance and accuracy by achieving the highest F1 and mAP values of 0.811 and 0.852 which are $6.61\%$ and $7.63\%$ improvements over the TPH-YOLOv5, respectively.
\begin{table}
\centering
\caption{Overall performance comparison  between TPH-YOLOv5, Dense-YOLOv4, and DenseSPH-YOLOv5.}
\begin{tabular}{c c c c  c c c }
%\hline
\\[-0.5em]
\hline
\\[-0.8em] % Adds extra space after hline
Detection model  &IoU  & F1  & mAP & Validation  loss   & \vtop{\hbox{\strut Detection }\hbox{time (ms) }} & \vtop{\hbox{\strut Detection }\hbox{speed (FPS) }} 
\\[-0.0em]
\hline
\\[-0.5em]
TPH-YOLOv5 & 0.740 & 0.745 & 0.776 & 18.07 & 22.47 & 44.5
\\
\\[-0.5em]
Dense-YOLOv4 & 0.781 & 0.782 &  0.819 & 10.13 & 19.30 & 51.8
\\
\\[-0.5em]
\bf{DenseSPH-YOLOv5} & 0.803 & 0.811  & 0.852 & 7.18  & 16.02 & 62.4
\\
\\[-0.5em]
\hline
\end{tabular}
\label{T-5}
\end{table}
\begin{figure}
\noindent
\centering
\includegraphics[width=1.1\linewidth]{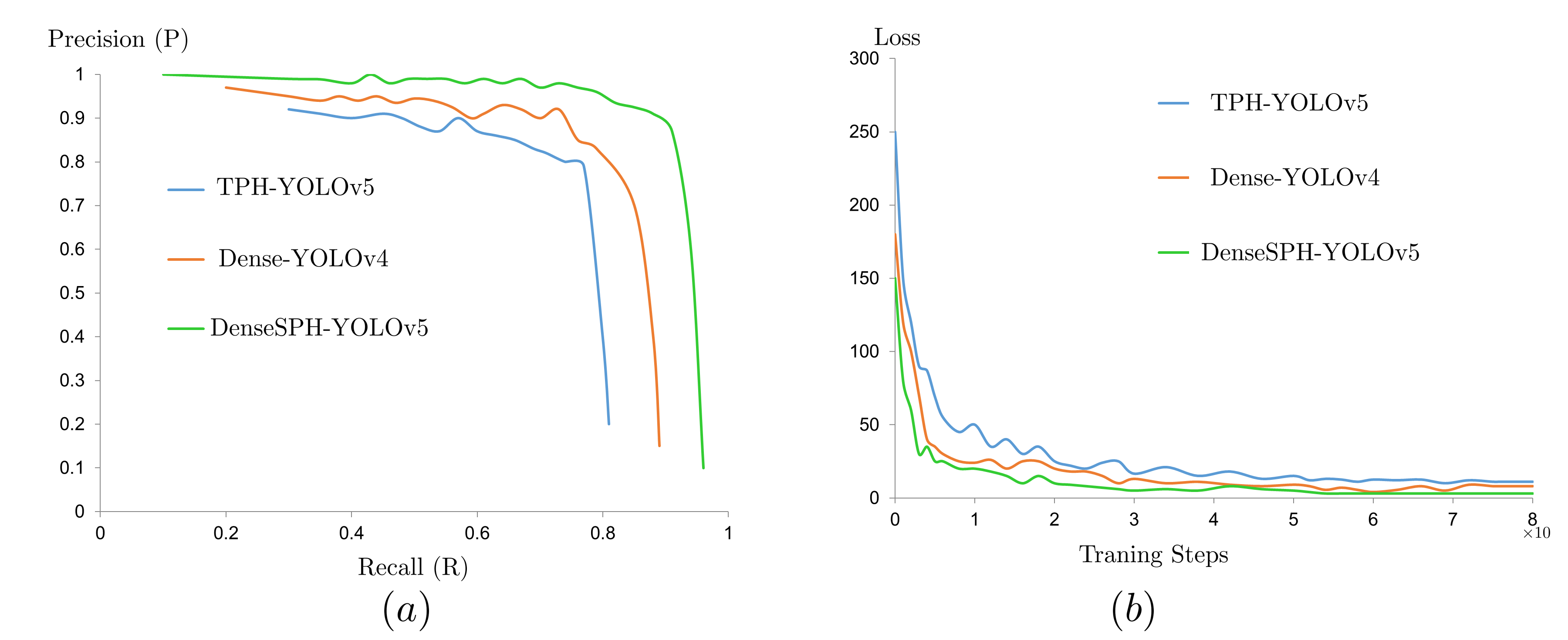}
\caption{\label{Fig-7} Comparison of (a) P-R curves; (b)  loss evolution curves  between  TPH-YOLOv5, Dense-YOLOv4, and DenseSPH-YOLOv5.}
\end{figure}
Furthermore, the detection speed of 62.4 FPS obtained from DenseSPH-YOLOv5 was found to be higher than both  TPH-YOLOv5 and Dense-YOLOv4. Thus, it can provide real-time road damage detection with better accuracy compared to the other two state-of-the-art models. 
In addition, the comparison of P-R curves between the three models have been depicted in Fig \ref{Fig-7}-(a).
From the comparison of the P-R curves, one can see that DenseSPH-YOLOv5 attains a better P value for a particular R.
It achieved the highest area under the P-R curve indicating superior detection performance compared to TPH-YOLOv5 and Dense-YOLOv4. 
Next, we compare the loss evolution curves as shown in Fig \ref{Fig-7}-(b).
In the initial phase, after exhibiting several cycles of fluctuation,
the  loss in the DenseSPH-YOLOv5 model tends to saturate
after approximately 50 training steps with a final loss value of 7.18.
Whereas, the other two models exhibit higher fluctuation in loss evolution and yield higher final loss value. 
Evidently, the proposed TPH-YOLOv5
is easier to train with faster convergence characteristics demonstrating its efficacy from the computational point of view. 
\begin{table}[]
\centering
\caption{Comparison of detection results for individual classes between  TPH-YOLOv5, Dense-YOLOv4, and DenseSPH-YOLOv5}
\label{T-6}
\begin{tabular}{c c c c c  c c c c c c}
%\hline
\\[-0.5em]
\hline
\\[-0.8em] % Adds extra space after hline 
Model  & Class $\rightarrow$   & \bf{D00} & \bf{D01}  & \bf{D10} & \bf{D11} & \bf{D20} & \bf{D40} & \bf{D43}  & \bf{D44}  & \bf{Avg.} \\ \hline
\multirow{5}{*}{\vtop{\hbox{\strut TPH-YOLOv5    }\hbox{\strut\,\, }}} 
&P &0.87 & 0.62 &0.84 &0.87 &0.84 &0.87 & 0.87 & 0.81  & 0.82 
\\ \cline{2-11} 
&R &0.61 & 0.89 &0.49 &0.51 &0.71 &0.74 & 0.81 & 0.88  & 0.71 
\\ \cline{2-11} 
&F1 &0.72 & 0.73 &0.62 &0.64 &0.77 &0.79 & 0.83 & 0.84  & 0.75 
\\ \cline{2-11} 
& mAP &0.81 & 0.79 &0.78 &0.82 &0.73 &0.78 & 0.76 & 0.74  & 0.77 
\\ \hline
\multirow{5}{*}{\vtop{\hbox{\strut Dense-YOLOv4 }\hbox{\strut\,\, }}} 
& P &0.89 & 0.69 &0.91 &0.92 &0.89 &0.91 & 0.89 & 0.84  & 0.87
\\ \cline{2-11} 
& R &0.65 & 0.93 &0.52 &0.51 &0.78 &0.78 & 0.86 & 0.87  & 0.74
\\ \cline{2-11} 
&F1 &0.75 & 0.79 &0.66 &0.65 &0.83 &0.84 & 0.87 & 0.85  & 0.78
\\ \cline{2-11} 
& mAP &0.82 & 0.81 &0.84 &0.89 &0.77 &0.82 & 0.81 & 0.79 & 0.82
\\ \hline
\multirow{5}{*}{\vtop{\hbox{\strut \bf{DenseSPH-YOLOv5} }\hbox{\strut\,\, }}} 
& P &0.92 & 0.74 &0.93 &0.94 &0.93 &0.92 & 0.91 & 0.87 & \bf{0.89}
\\ \cline{2-11} 
& R &0.68 & 0.95 &0.58 &0.54 &0.81 &0.79 & 0.88 & 0.87 & \bf{0.76}
\\ \cline{2-11} 
& F1 &0.78 & 0.83 &0.71 &0.68 &0.86 &0.85 & 0.89 & 0.87 & \bf{0.81}
\\ \cline{2-11} 
& mAP &0.84 & 0.89 &0.81 &0.91 &0.82 &0.85 & 0.85 & 0.85 & \bf{0.85}
\\ \hline
\end{tabular}
\end{table}

\begin{figure}
\centering
\includegraphics[width=\textwidth]{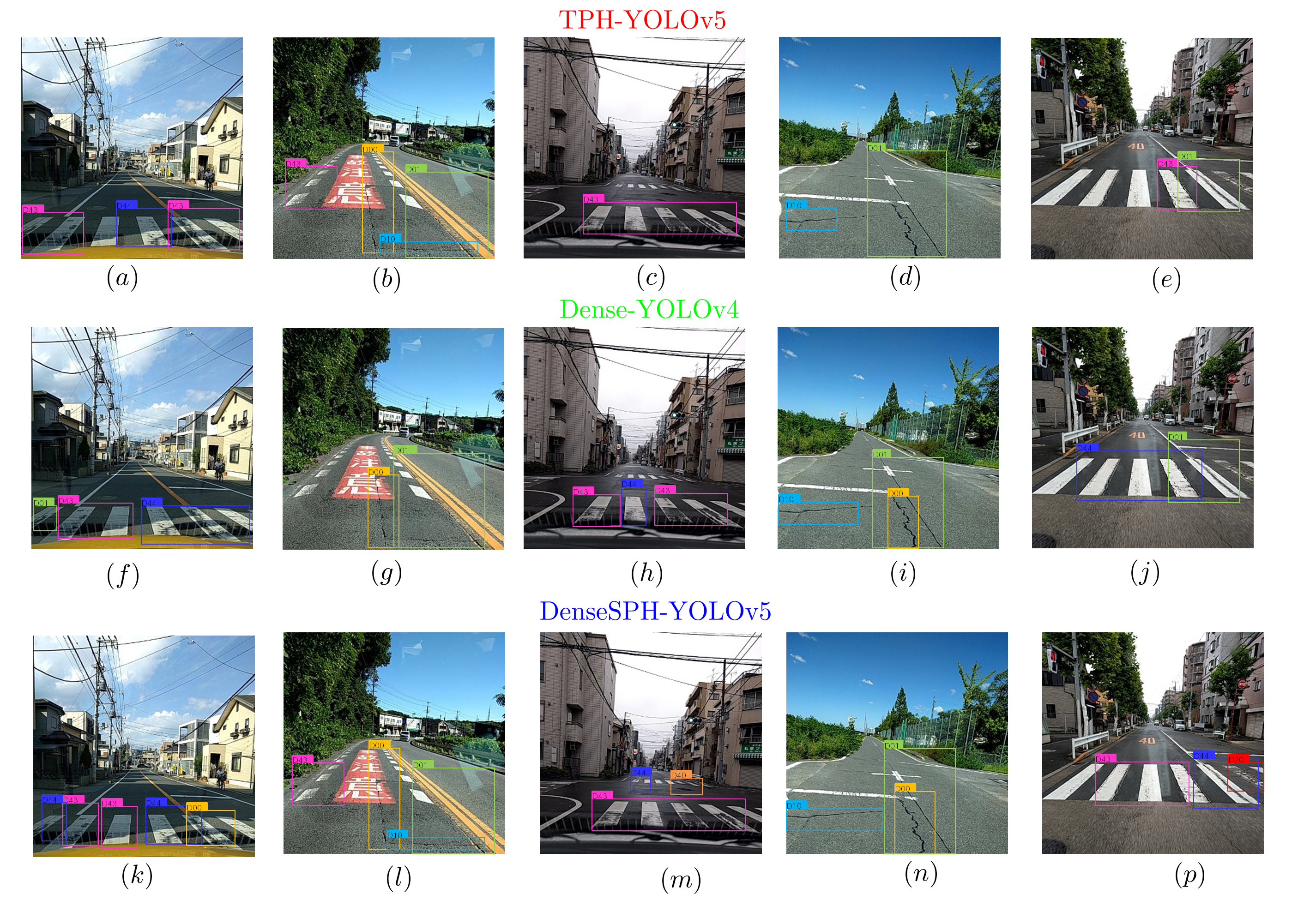}
\caption{Comparison of various damage detection results  from (a-e) TPH-YOLOv5; (f-j) Dense-YOLOv4; (k-p) DenseSPH-YOLOv5.  Detailed detection results with average confidence scores have been shown in Table \ref{T-7}.}
\label{Fig-9}
\end{figure}

\begin{figure}
\centering
\includegraphics[width=0.95\textwidth]{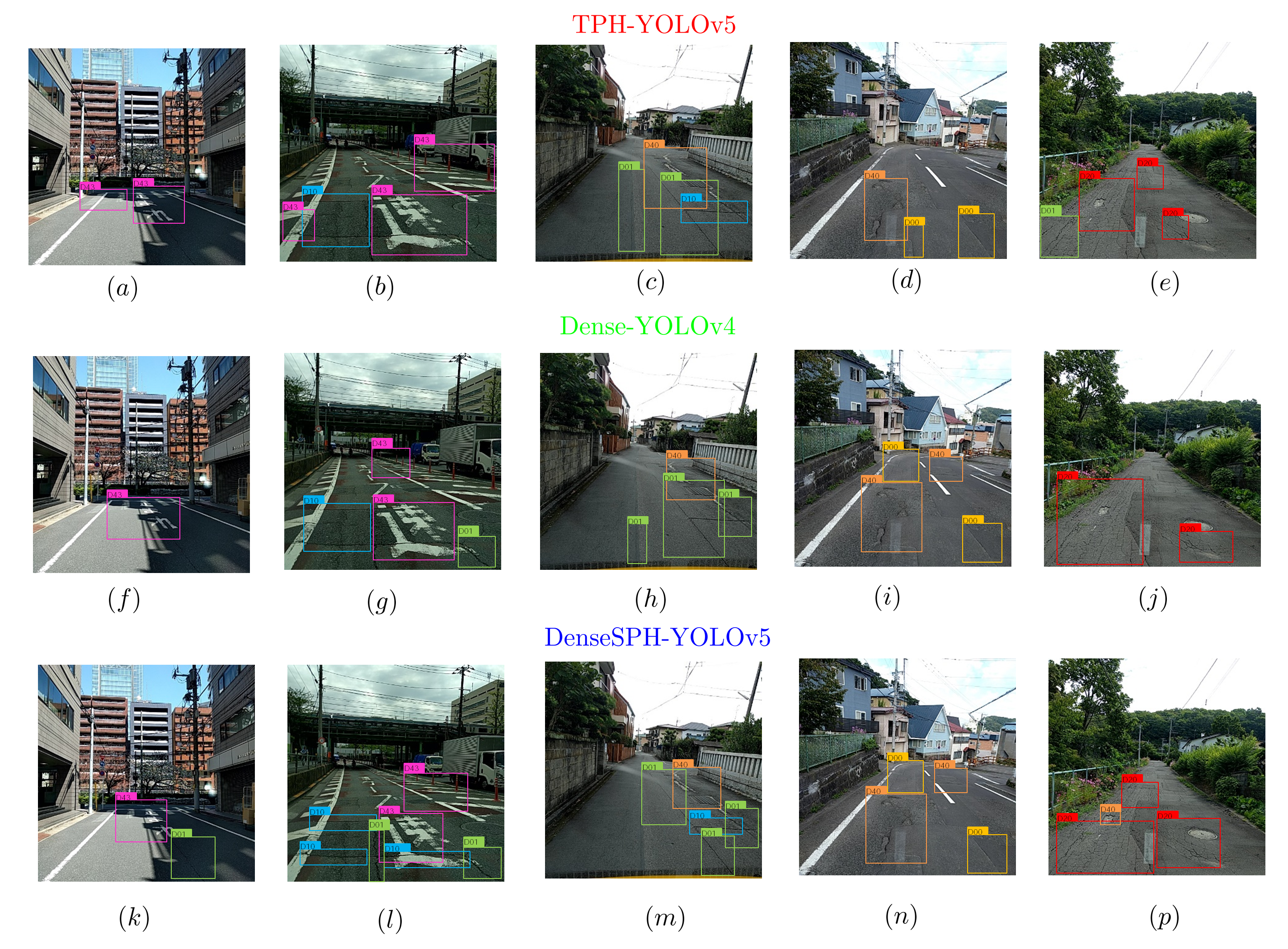}
\caption{Comparison of damage detection results under various challenging scenarios from (a-e) TPH-YOLOv5; (f-j) Dense-YOLOv4; (k-p) DenseSPH-YOLOv5.  Detailed detection results with average confidence indexes have been shown in Table \ref{T-6}.}
\label{Fig-10}
\end{figure}

\begin{figure}
\centering
\includegraphics[width=0.95\textwidth]{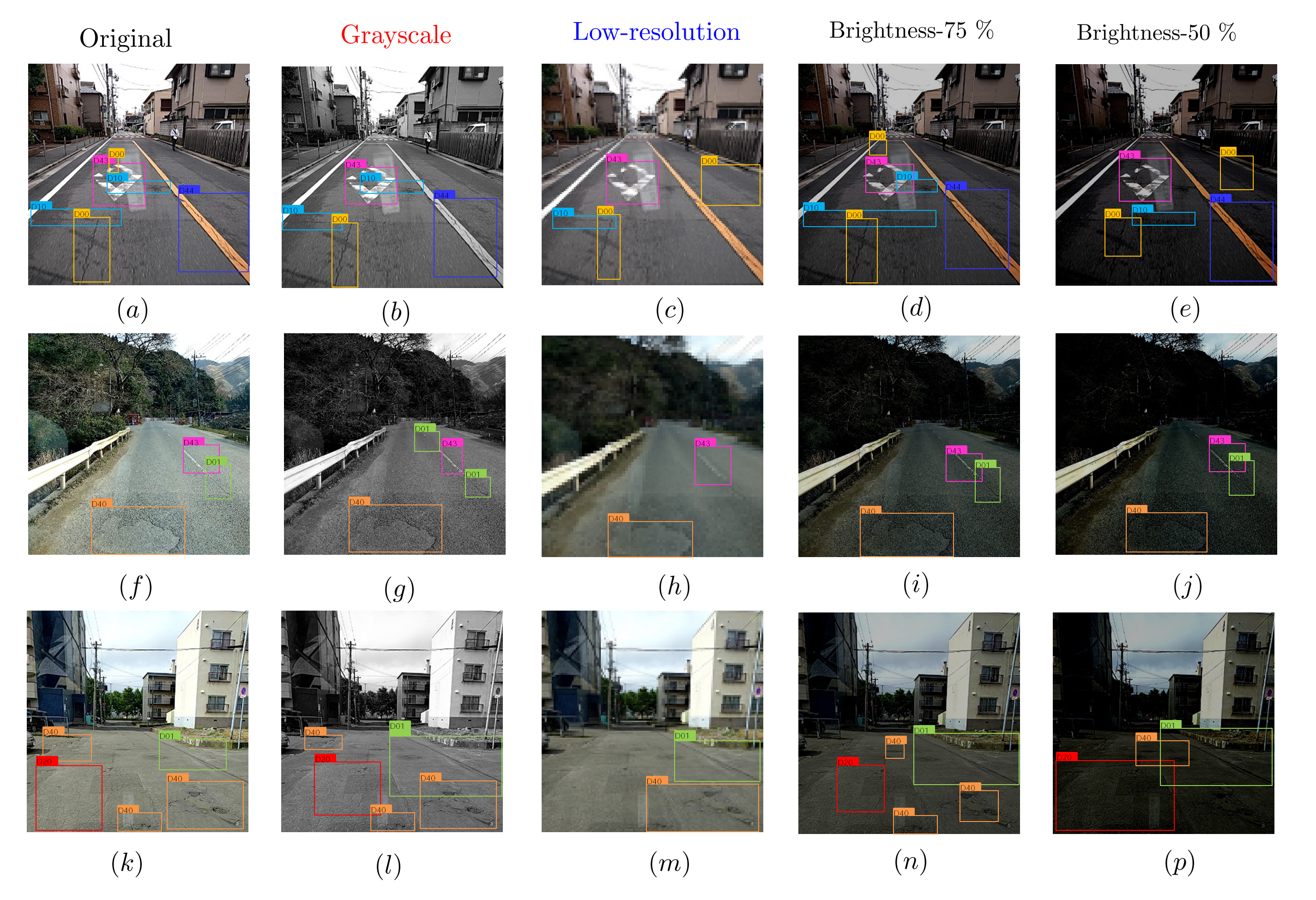}
\caption{Detection results for (a, f, k) original RGB images; (b, g, l) corresponding grey scale images; (c, h, m)  low resolution images; (d, i, n) 75 \% brightness intensity; (e, j, p) 50 \% brightness intensity   from the DenseSPH-YOLOv5.  Detailed detection results have been shown in Table \ref{T-6}.}
\label{Fig-11}
\end{figure}

To further gain insight into the performances of these models, a comparison of detection results containing  P, R, mAP, and F-1 values from each individual class between  TPH-YOLOv5, Dense-YOLOv4, and DenseSPH-YOLOv5 has been presented in Table \ref{T-6}. 
DenseSPH-YOLOv5 has illustrated significant improvement in P and R values for various classes, in particular, for detecting longitudianl and lateral linear cracks, alligator cracks, and potholes.
Thus,  DenseSPH-YOLOv5 efficiently maximizes the TP value while simultaneously reducing FP and FN values for all classes.
The proposed model improves $2.01\%$ in P and $2.11\%$ in R compared to Dense-YOLOv4. 
From the overall comparison, we can conclude that DenseSPH-YOLOv5 demonstrated the best performance in detecting various damage classes outperforming the other two state-of-the-art models in terms of precision and accuracy.
\begin{figure}
\centering
\includegraphics[width=0.95\textwidth]{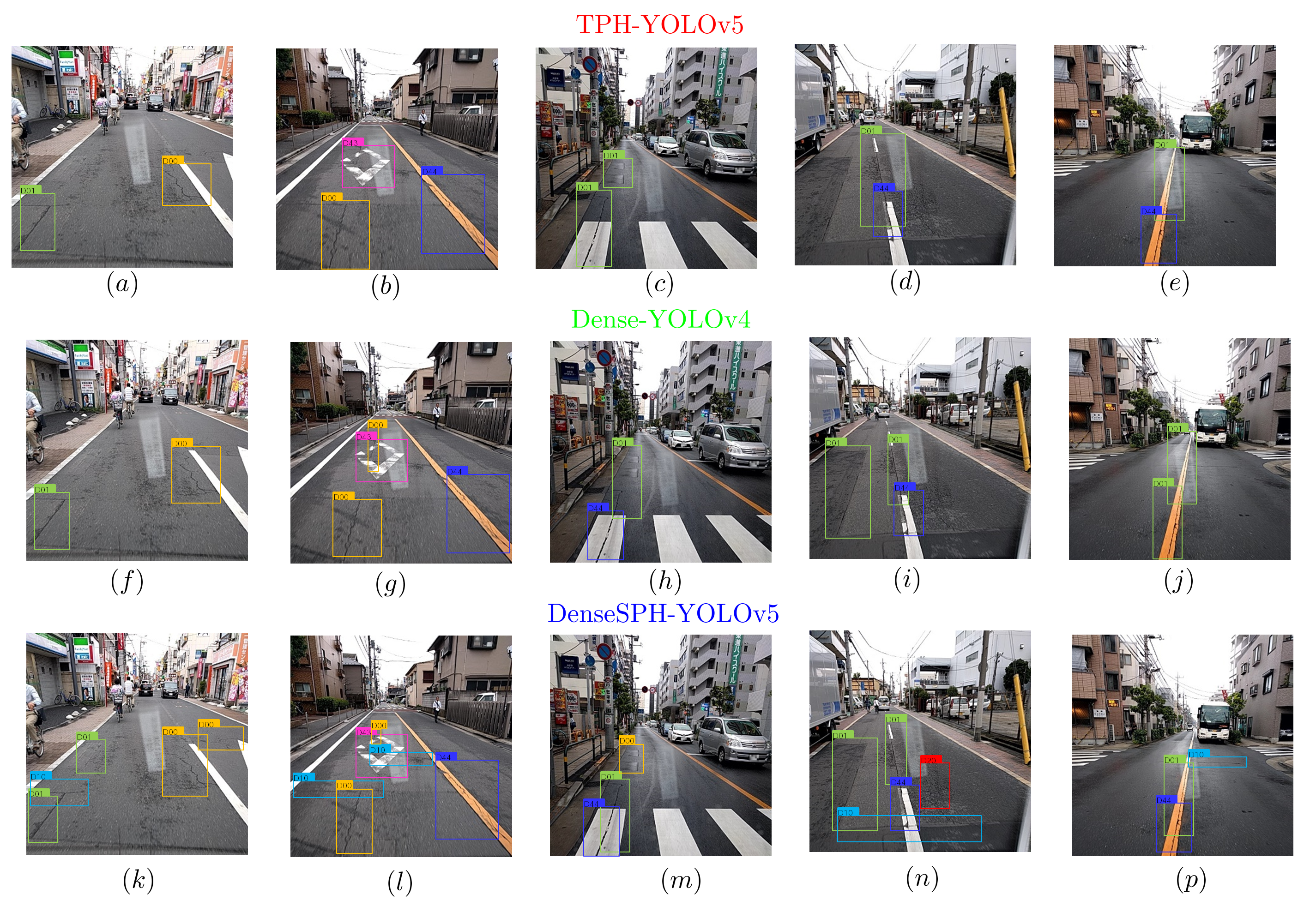}
\caption{Comparison of various damage detection results  from (a-e) TPH-YOLOv5; (f-j) Dense-YOLOv4; (k-p) DenseSPH-YOLOv5.  Detailed detection results with average confidence indexes have been shown in Table \ref{T-7}.}
\label{Fig-8}
\end{figure}
\begin{table}
\centering
\caption{Detailed detection results from  TPH-YOLOv5, Dense-YOLOv4, and DenseSPH-YOLOv5 for different damage classes as shown in Figs.  \ref{Fig-8}- \ref{Fig-9}. }
\begin{tabular}{c c c c c c c}
%\hline
\\[-0.5em]
\hline
\\[-0.8em] % Adds extra space after hline
Figs. No & Model  &  Detc.   & False/Undetc.   & Avg. confidence Score 
\\[-0.0em]
\hline
\\[-0.5em]
\ref{Fig-8} (a-e) & TPH-YOLOv5   & 13 & 8  & 0.71
\\
\\[-0.5em]
\ref{Fig-8} (f-j) & Dense-YOLOv4    & 16 & 5  & 0.75
\\
\\[-0.5em]
\ref{Fig-8} (k-p) & \bf{DenseSPH-YOLOv5}    & 20 & 1  & 0.83
\\
\\[-0.5em]
\hline     
\\[-0.5em]
\ref{Fig-9} (a-e) & TPH-YOLOv5    & 13 & 7  & 0.67
\\
\\[-0.5em]
\ref{Fig-9} (f-j) & Dense-YOLOv4   & 14 & 6  & 0.72
\\
\\[-0.5em]
\ref{Fig-9} (k-p) & \bf{DenseSPH-YOLOv5}    & 18 & 2  & 0.79
\\
\\[-0.5em]
\hline     
\end{tabular}
%}
\label{T-7}
\end{table}
\\
\\
{\bf 4.6 Detection of various  damage classes: }
\\
\\
In this section, we have demonstrated the detection results obtained from DenseSPH-YOLOv5  for eight different damage classes and compared them with TPH-YOLOv5 and  Dense-YOLOv4. The visual representations of the detection results have been presented with confined bounding boxes with class identifiers (see Table \ref{T-1})  as shown in Figs. \ref{Fig-8}- \ref{Fig-9}. 
Corresponding detailed detection results consisting of the number of detected and undetected target classes with average confidence scores have been reported in Table. \ref{T-7}. 
From the overall comparison, it can be concluded that the proposed model shows its efficacy by preciously detecting the target objects with high average confidence index values. At the same time, it minimizes the false and missed-detection results compared to both TPH-YOLOv5 and  Dense-YOLOv4 models. Particularly, when the multiple target objects have a significant degree of overlap between them, the bounding box prediction from the proposed DenseSPH-YOLOv5 is quite accurate in detecting each target object as illustrated in Figs. \ref{Fig-8}-(l, n) and Figs. \ref{Fig-9}-(k, n). 
\\
\\
{\bf 4.7 Detection under challenging scenarios:}
\\
\\
In this section, we have extended the detection under some challenging scenarios as shown in  Fig. \ref{Fig-10} to demonstrate the efficacy of the proposed DenseSPH-YOLOv5. 
The detailed detection results that consist of the total number of detected and undetected target classes with average confidence scores have been reported in Table. \ref{T-8}.
\begin{table}
\centering
\caption{Detailed detection results from  TPH-YOLOv5, Dense-YOLOv4, and DenseSPH-YOLOv5 for different damage classes as shown in Fig.  \ref{Fig-10}. }
\begin{tabular}{c c c c c c c  }
%\hline
\\[-0.5em]
\hline
\\[-0.8em] % Adds extra space after hline
Figs. No & Model  &  Detc.   & False/Undetc.   & Avg. confidence Score 
\\[-0.0em]
\hline
\\[-0.5em]
\ref{Fig-10} (a)-(e) & TPH-YOLOv5    & 13 & 10  & 0.68
\\
\\[-0.5em]
\ref{Fig-10} (f)-(j) & Dense-YOLOv4    & 15 & 8  & 0.69
\\
\\[-0.5em]
\ref{Fig-10} (k)-(p) & \bf{DenseSPH-YOLOv5}    & 19 & 4  & 0.77	
\\
\\[-0.5em]
\hline     
\end{tabular}
%}
\label{T-8}
\end{table}
Here we consider various complex backgrounds and challenging environments where the target class is predominately difficult to detect due to the presence of shadows. Additionally, there is some degree of occlusion of overlapping bounding boxes between target classes.
It is a challenging task to detect target objects individually for an object detection model.
%similar background texture, high degree of occlusion, and dense overlapping between object classes.
Thus, superior feature extraction is critical to detect various target classes efficiently and correctly.  In such cases, the proposed DenseSPH-YOLOv5 has demonstrated its superiority in terms of boundary box precision with a higher average confidence score compared to the other two models. Notably, it significantly reduces the false or missed detection (10 to 4 compared to TPH-YOLOv5) and consequently, the confidence scores have been improved which is evident from Table \ref{T-8}. 
Overall, based on the detection results for the various damage classes, the proposed DenseSPH-YOLOv5 illustrates superior detection ability, in particular, in detecting the presence of shadows compared to TPH-YOLOv5 and Dense-YOLOv4  models. 
\\
\\
{\bf 4.8 Detection under Greyscale,  low resolution, and different illumination intensities:}
\\
\\
In this section, the detection accuracy of the DenseSPH-YOLOv5  has been further tested with grayscale and pixelated low-resolution $(100\times100)$ images as shown in Fig. \ref{Fig-11}. 
Additionally,  prediction results under different illumination intensities (i.e., brightness-$75\%$  and brightness- $50\%$) have been studied and compared with the detection result for original RGB images. 
In Table \ref{T-9}, detailed detection results with confidence scores have been reported.  
\begin{table}
\centering
\caption{ Detection result comparison from the DenseSPH-YOLOv5 for original RGB, greyscale,  low resolution, and various brightness intensity  (75 \% and 50 \%) images  as shown in Fig. \ref{Fig-11}.}
\begin{tabular}{c*{12}{c}}%\toprule
\hline
%\multirow{2}
{ Figs. No} & \multicolumn{2}{c}{ Original } & \multicolumn{2}{c}{ Grey scale } &
\multicolumn{2}{c}{ Low resolution } &
\multicolumn{2}{c}{ Brightness-75 \%} &
\multicolumn{2}{c}{ Brightness-50 \% } \\
%\cmidrule(lr){3-4}\cmidrule(lr){5-6} \cmidrule(lr){7-8}
&  Det.   & Undet.  & Det.   & Undet. & Det.   & Undet. &  Det.   & Undet.  & Det.   & Undet. \\ \hline  \\%\midrule
Figs. \ref{Fig-11}-(a-e)   & 6   & 0    & 5    & 1   & 4   & 2     & 6   & 0   & 5     & 1     \\
Figs. \ref{Fig-11}-(f-j)   & 3   & 1    & 4    & 0    & 2   & 2    & 3  & 1  & 3    & 1     \\
Figs. \ref{Fig-11}-(k-p)   & 5   & 0   & 5    & 0    & 2   & 3     & 5   & 0   & 3     & 2     
\\ \\\hline %\bottomrule
\end{tabular}
\label{T-9}
\end{table}
For grayscale images, the proposed model has demonstrated impressive performance with accurate bounding box prediction with a better bounding box confidence score as shown in Figs. \ref{Fig-11}-(b, g, l).
For pixelated low-resolution images, the performance of DenseSPH-YOLOv5 has been slightly diminished, however, it can still detect most of the target class with reasonable accuracy. 
Similarly,  the proposed model demonstrates superior detection results under different illumination intensities as demonstrated in Figs. \ref{Fig-11}-(d, i, n) and Figs. \ref{Fig-11}-(e, j, p).
It can be concluded from the test results that the proposed model is more adaptive in more challenging environments compared to  TPH-YOLOv5 and Dense-YOLOv4  models.
\\
\\
{\bf 5. Discussion:}
\\
\\
From the overall detection result, it is evident that DenseSPH-YOLOv5 has higher adaptability and better capability in localizing various damage classes in a complex environment and challenging conditions compared to current state-of-the-art models. 
Additionally,  it has demonstrated higher accuracy in bounding box detection and therefore can effectively avoid the problem of
false and missed detection over other detection networks which justifies the usefulness of the proposed model in practical on-field damage detection tasks. Future work can be geared towards 
further optimization of detection accuracy as well as detection speed for portable on-field damage detection in a mobile computing platform. 
The current model can be further improved by incorporating some state-of-the-art algorithms such as   YOLO-X \citep{ge2021yolox}, YOLOv7 \citep{wang2022yolov7},  etc. 
Additionally, it can be assembled with a drone video surveillance system for real-time accurate damage detection. 
Furthermore, overlapping and occultation can lead to the wrong prediction which can be further improved by considering shortening the feature stride for better localization and bounding box prediction. 
In order to avoid overlapping boxes, it can be useful to switch from bounding boxes to points \citep{ribera2019locating} or
masks \citep{xu2020automated}. 
Moreover, one of the potential applications can be assembling object detection framework with semantic segmentation methods such as Mask R-CNN \citep{bharati2020deep}, U-Net \citep{esser2018variational} to extract morphological information of various damage classes. 
Nonetheless, the current work elucidates the possibility of applying cutting-edge DL-based computer vision methodology for multiclass automated damage detection processes for commercial applications.
\\
\\
{\bf 6.  Conclusions :}
\\
\\
Summarizing, in the present work, we have developed an efficient and robust object localization model 
DenseSPH-YOLOv5 based on an improved version of the YOLOv5 network for accurate classification and localization of damage detection. 
The proposed DenseSPH-YOLOv5 has been employed to detect distinct eight different road damage classes that provide superior and accurate detection under various complex and challenging environments.  
Evaluated on the RDD-2018 dataset, it has been found that at a detection rate of 62.40 FPS,   DenseSPH-YOLOv5 has achieved mean-average precision (mAP), F1, and precision (P) values of  $85.25 \%$,  $81.18 \%$, and  $89.51 \%$, respectively outperforms existing state-of-the-art damage detection models in terms of both classification accuracy and localized bounding box prediction in detecting all damage classes.
The present work effectively addresses the shortcoming of existing DL-based damage detection models and illustrates its superior potential in real-time in-field applications. 
In short, current work constitutes a step toward a  fully automated and efficient  DL-based computer vision methodology for multi-class damage detection processes. 
%% \\
%\\
%{\bf Author contributions :}
%\\
%\\
%Arunabha Mohan Roy:   Conceptualization; Data curation; Formal analysis; Investigation; Methodology; Software; Validation; Visualization; Roles/Writing - original draft; Writing - review, and editing.
%:  Formal analysis; Investigation; Methodology; Roles/Writing - original draft ; Writing - review, and  editing; Supervision; Project Administration.
\\
\\
{\bf Acknowledgements:}
The support of the Aeronautical Research and Development Board (Grant No.
DARO/08/1051450/M/I) is gratefully
acknowledged.
\\
\\
{\bf Conflict of  interest:}
The authors declare that they have no known competing financial interests or personal relationships that could have appeared to influence the work reported in this paper.
%\\
%\\
%{\bf Data availability:} The data that support the findings of this study are publicly available from \url{uuu} 

%	%========================================================
%	%for bibtex style  use this (if you dont use "natbib") %
%	%plain,%unsrt, %abbrv,spphys, %acm,%authordate1, %apacite,%named
%	
%	% \bibliographystyle{plain}
%	%\bibliographystyle{acm}
%	\bibliographystyle{elsarticle-num}
%	
%	
%	%SPRINGER%
%	%\bibliographystyle{unsrtnat}  
%	%\bibliographystyle{spbasic}      % basic style, author-year citations
%	%\bibliographystyle{spmpsci}      % mathematics and physical sciences
%	%\bibliographystyle{spphys}       % APS-like style for physics
%	%==============================================
%	%with "natbib" (example refrence style)
%	%==============================================
%	
%	
% %==============================================
\newpage
\bibliographystyle{apalike}

\bibliography{Ref-YOLOv5} 

%==============================================	

\end{document}